\documentclass[10pt,twocolumn,letterpaper]{article}

\usepackage{iccv}
\usepackage{times}
\usepackage{epsfig}
\usepackage{graphicx}
\usepackage{amsmath}
\usepackage{amssymb}
\usepackage{multirow}
\usepackage{tabularx}
\usepackage{caption}
\usepackage{subcaption}
\usepackage{makecell}

\usepackage{pifont}
\newcommand{\xmark}{\ding{55}}%
\newcommand{\indep}{\perp \!\!\! \perp}

\usepackage[pagebackref=true,breaklinks=true,letterpaper=true,colorlinks,bookmarks=false]{hyperref}

\iccvfinalcopy 

\def\iccvPaperID{1961} 

\ificcvfinal\fi

\begin{document}

\title{Evidential Deep Learning for Open Set Action Recognition}

\author{Wentao Bao, Qi Yu, Yu Kong\\
Rochester Institute of Technology, Rochester, NY 14623, USA\\
{\tt\small \{wb6219, qi.yu, yu.kong\}@rit.edu}
}

\maketitle
\ificcvfinal\thispagestyle{empty}\fi

\begin{abstract}
   In a real-world scenario, human actions are typically out of the distribution from training data, which requires a model to both recognize the known actions and reject the unknown. Different from image data, video actions are more challenging to be recognized in an open-set setting due to the uncertain temporal dynamics and static bias of human actions. In this paper, we propose a Deep Evidential Action Recognition (DEAR) method to recognize actions in an open testing set. Specifically, we formulate the action recognition problem from the evidential deep learning (EDL) perspective and propose a novel model calibration method to regularize the EDL training. Besides, to mitigate the static bias of video representation, we propose a plug-and-play module to debias the learned representation through contrastive learning. Experimental results show that our DEAR method achieves consistent performance gain on multiple mainstream action recognition models and benchmarks. Code and pre-trained models are available at {\small{\url{https://www.rit.edu/actionlab/dear}}}.
\end{abstract}

\section{Introduction}

Video action recognition aims to classify a video that contains a human action into one of the pre-defined action categories (closed set). However, in a real-world scenario, it is essentially an \textit{open set} problem~\cite{ShuICME2018}, which requires the classifier to simultaneously recognize actions from known classes and identify actions from unknown ones~\cite{ScheirerTPAMI2012,GengTPAMI2020}. In practice, open set recognition (OSR) is more challenging than closed set recognition, while it is important for applications such as face recognition~\cite{LiuCVPR2017}, e-commerce product classification~\cite{XuWWW2019}, autonomous driving~\cite{RoitbergIV2020}, and so on. 

\begin{figure}
    \centering
    \subcaptionbox{Kinetics~\cite{kinetics}}{
    \includegraphics[width=0.47\linewidth]{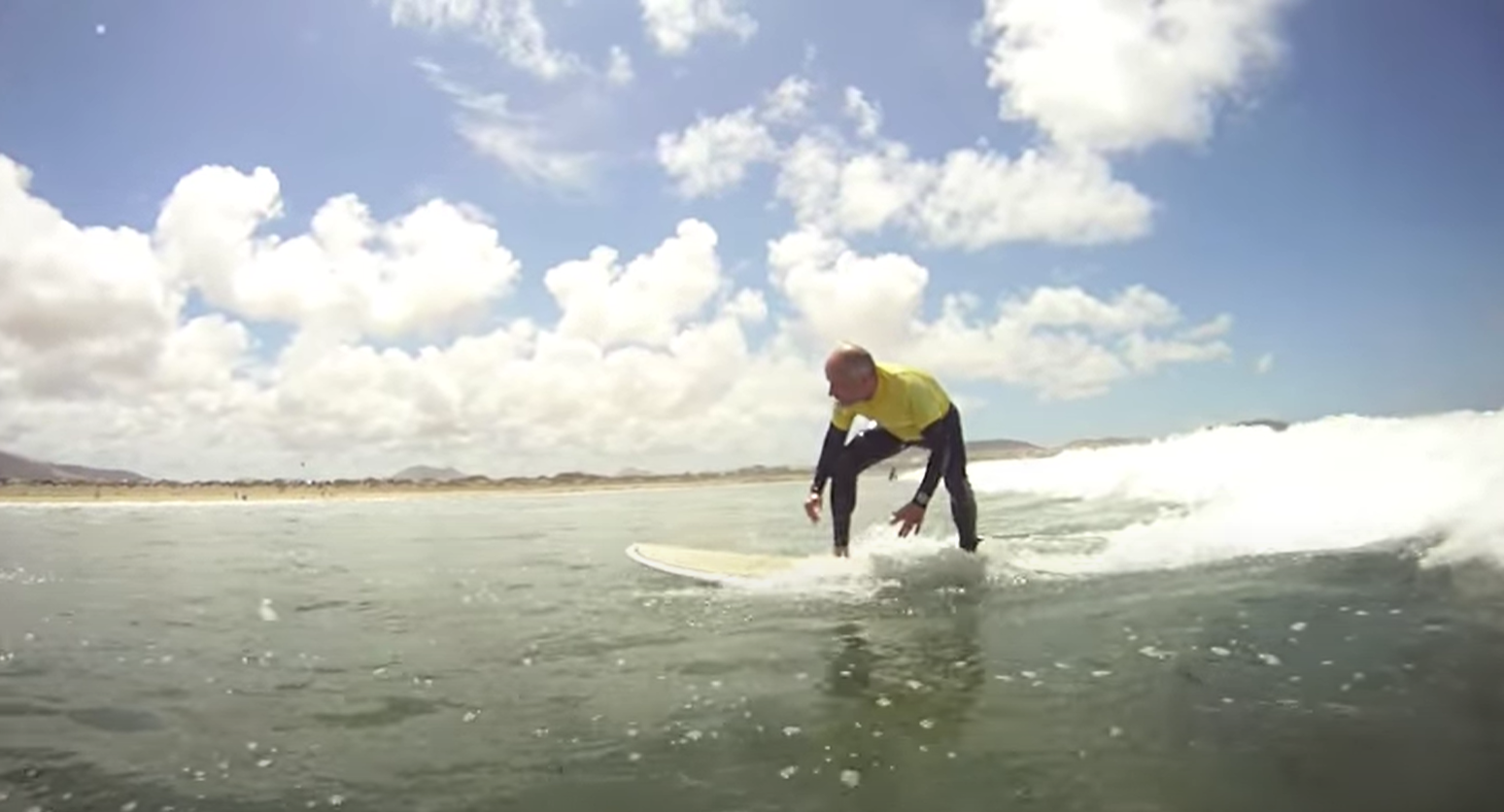}
    }
    \subcaptionbox{Mimetics~\cite{mimetics}}{
    \includegraphics[width=0.47\linewidth]{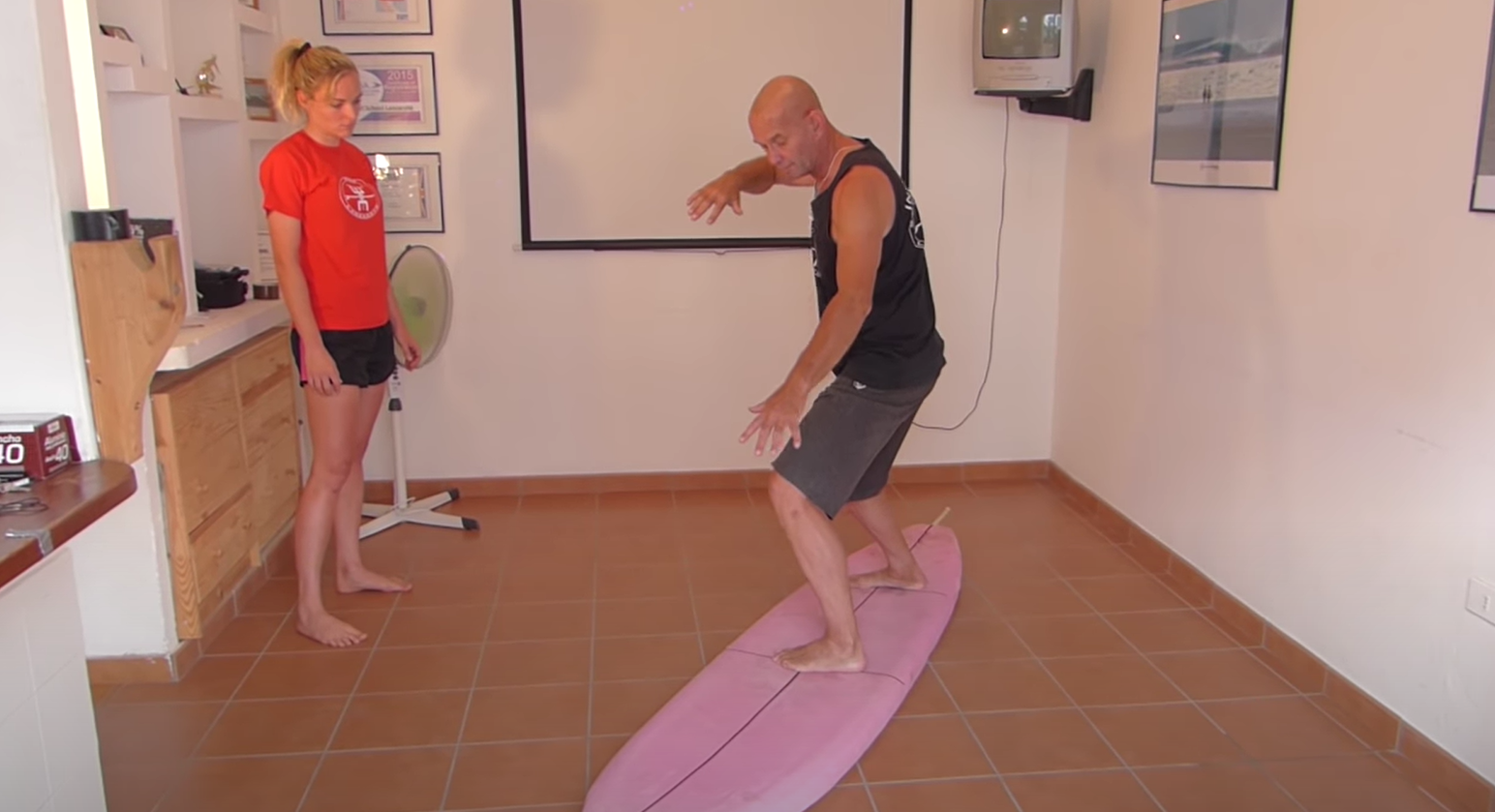}
    }
    \captionsetup{font=small,aboveskip=5pt}
    \caption{\textbf{An Example of Static Bias.} To recognize the human action (i.e., ``Surfing Water"), the recognition model which is biased to the background of water and sky in the closed set (Kinetics) would be unable to recognize the same action with the indoor scene in open set (Mimetics as unknown).}
    \label{fig:bias}
\end{figure}

OSR was originally formalized in~\cite{ScheirerTPAMI2012} and many existing approaches have been proposed using image datasets such as MNIST~\cite{mnist} and CIFAR-10~\cite{cifar10}. However, unlike OSR, limited progress has been achieved for open set action recognition (OSAR) which is increasingly valuable in practice. In fact, novel challenges arise in OSAR 
from the following key aspects. First, the temporal nature of videos may lead to a high diversity of human action patterns. Hence, an OSAR model needs to capture the temporal regularities of closed set actions but also be \textit{aware of what it does not know} when presented with unknown actions from an open set scenario. Second, the visual appearance of natural videos typically contain \textit{static biased} cues~\cite{LiECCV2018,ChoiNIPS2019} (e.g., ``surfing water" in totally different scenes as shown in Fig.~\ref{fig:bias}). Without addressing the temporal dynamics of human actions, the static bias could seriously hamper the capability of an OSAR model to recognize unknown actions from an unbiased open set. Due to these challenges, existing effort on OSAR is quite limited with few exceptions~\cite{ShuICME2018,KrishnanNIPS2018,YangPR2019}. They simply regard each video as a standalone sample and primarily rely on image-based OSR approaches. As a result, they fall short in addressing the inherent video-specific challenges in the open set context as outlined above.

\begin{figure}
    \centering
    \includegraphics[width=0.49\linewidth]{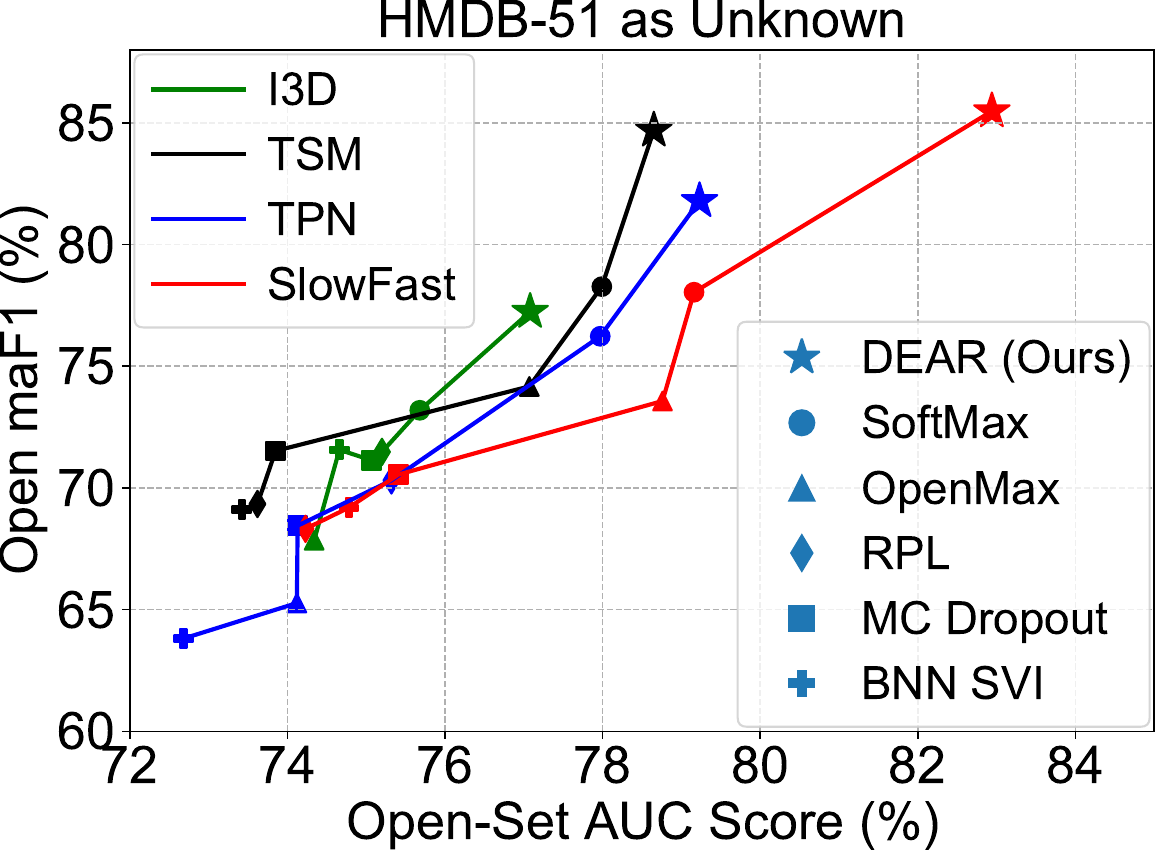}
    \includegraphics[width=0.49\linewidth]{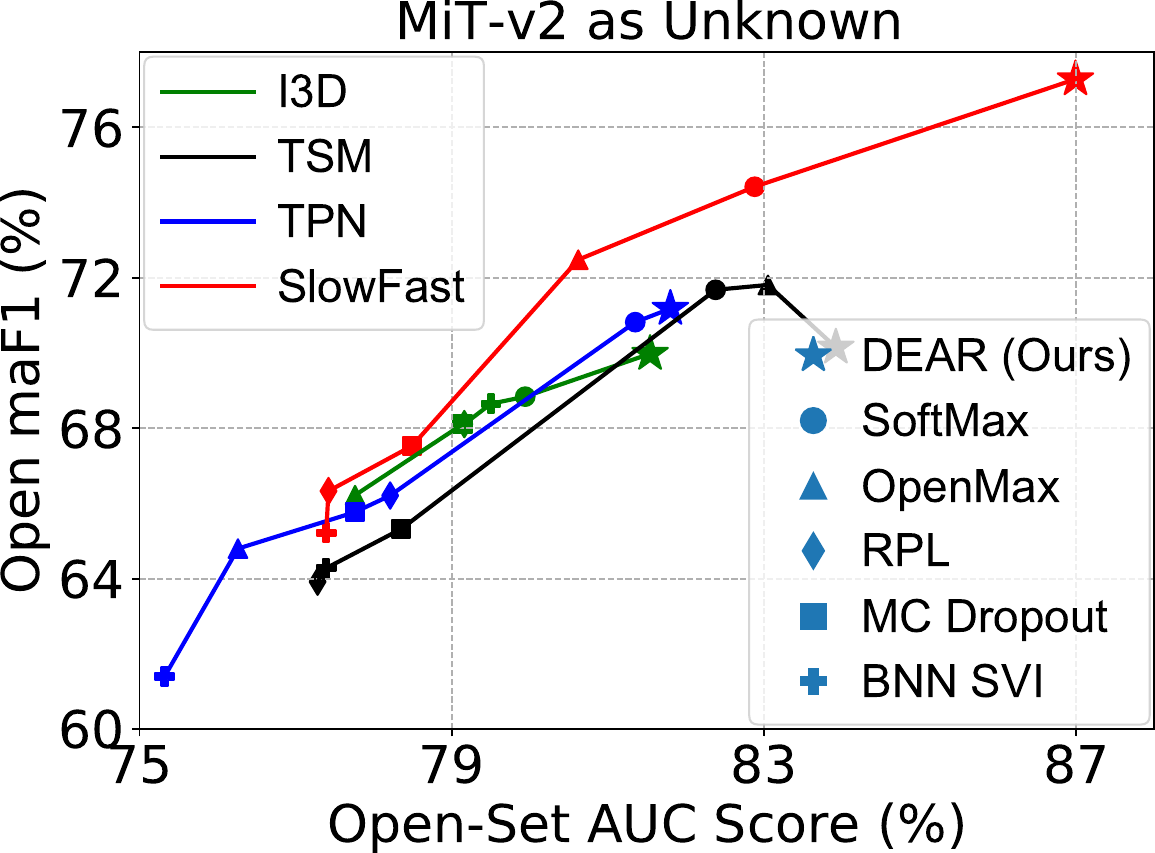}
    \captionsetup{font=small,aboveskip=5pt}
    \caption{\textbf{Open Set Action Recognition Performance.} HMDB-51~\cite{hmdb51} and MiT-v2~\cite{mitv2} are separately used as small- and large-scale unknown data for models trained on the closed set UCF-101~\cite{ucf101}. Our DEAR method ($\star$) significantly outperforms existing approaches on multiple action recognition models. 
    }
    \label{fig:gain_curve}
\end{figure}

In this paper, we propose a Deep Evidential Action Recognition (DEAR) method for the open set action recognition task. To enable the model to ``\textit{know unknown}" in an OSAR task, our method formulates it as an uncertainty estimation problem by leveraging evidential deep learning (EDL)~\cite{SensoyNIPS2018,ZhaoAAAI2019,ShiNIPS2020,AminiNIPS2020,SensoyAAAI2020}. EDL utilizes deep neural networks to predict a Dirichlet distribution of class probabilities, which can be regarded as an evidence collection process. The learned evidence is informative to quantify the predictive uncertainty of diverse human actions so that unknown actions would incur high uncertainty, i.e., the model knows the unknown. Furthermore, to overcome the potential over-fitting risk of EDL in a closed set, we propose a novel model calibration method to regularize the evidential learning process. Besides, to mitigate the static bias problem for video actions, we propose a plug-and-play module to debias the learned representation through contrastive learning. Benefiting from the evidential theory, our DEAR method is practically flexible to implement and provides a principled way to quantify the uncertainty for identifying the unknown actions. Experimental results show that the DEAR method boosts the performance of existing powerful action recognition models with both small and large-scale unknown videos (see Fig.~\ref{fig:gain_curve}), while still maintains a high performance in traditional closed set recognition setting.

Distinct from existing OSR methods~\cite{ShuICME2018,KrishnanNIPS2018}, the proposed DEAR is the first evidential learning model for large-scale video action recognition. DEAR is superior to existing Bayesian uncertainty-based methods~\cite{KrishnanNIPS2018} in that model uncertainty can be directly inferred through evidence prediction that avoids inexact posterior approximation or time-consuming Monte Carlo sampling~\cite{AminiNIPS2020}. Moreover, our proposed model calibration method ensures DEAR to be confident in accurate predictions while being uncertain about inaccurate ones. Compared to~\cite{ShuICME2018} that incrementally learns a classifier for unknown classes, our method is more flexible in training without the access to unknown actions. Moreover, our proposed debiasing module could reduce the detrimental static bias of video actions so that the model is robust to out-of-context actions in the open set setting.

In summary, the contribution of this paper is three-fold:
\begin{itemize}
    \item Our Deep Evidential Action Recognition (DEAR) method performs novel evidential learning to support open set action recognition with principled and efficient uncertainty evaluation.
    \item The proposed Evidential Uncertainty Calibration (EUC) and Contrastive Evidential Debiasing (CED) modules effectively mitigate over-confident predictions and static bias problems, respectively.
    \item The DEAR method is extensively validated and consistently boosts the performance of state-of-the-art action recognition models on challenging benchmarks.
\end{itemize}

\section{Related Work}

\textbf{Open Set Recognition.} OSR problem originates from face recognition scenario~\cite{LiTPAMI2005} and it is firstly formalized by Scheirer~\etal~\cite{ScheirerTPAMI2012}. In~\cite{ScheirerTPAMI2012}, to reject the unknown classes, a binary support vector machine (SVM) was introduced by adding an extra hyper-plane for each new class. Based on this work, the Weibull-calibrated SVM (W-SVM)~\cite{ScheirerTPAMI2014} and $P_I$-SVM~\cite{JainECCV2014} are further proposed to calibrate the class confidence scores by leveraging the statistical extreme value theory (EVT). 
With the recent success of deep learning, deep neural networks (DNN) are widely used in OSR problem. To overcome the drawbacks of softmax in DNN, Bendale~\etal~\cite{BendaleCVPR2016} proposed OpenMax to bound the open space risk for DNN models. Based on this work, G-OpenMax~\cite{GeBMVC2017} adopted generative method to synthesize unknown samples in the training of DNNs. Similarly, recent deep generative adversarial networks (GANs) were used to generate samples of unknown class for OSR task~\cite{NealECCV2018,DitriaACCV2020}. To reject the unknown, variational auto-encoder (VAE) was recently used to learn the reconstruction error in OSR task~\cite{OzaCVPR2019,yoshihashiCVPR2019,SunCVPR2020}. 
Different from these methods, our method is the first work to introduce the evidential deep learning (EDL) for the OSR task and show the advantage over existing approaches.

For open set action recognition (OSAR) problem, it is much more challenging than OSR problem while only a few existing literature explored it. Shu~\etal~\cite{ShuICME2018} proposed ODN by incrementally adding new classes to the action recognition head. To capture the uncertainty of unknown classes, Bayesian deep learning is recently introduced to identify the unknown actions in~\cite{KrishnanNIPS2018,SubedarICCV2019,KrishnanAAAI2020}. Busto~\etal~\cite{BustoTPAMI2018} proposed an open set domain adaptation method. 
However, existing methods ignore the importance of uncertainty calibration and static bias of human actions in video data. In a broader context, uncertainty-based OSR is also closely related to out-of-distribution (OOD)~\cite{TranNIPS2020}. 
Other less related topics such as anomaly detection~\cite{PangArXiv2020}, generalized zero-shot learning~\cite{MandalCVPR2019}, and open world learning~\cite{BendaleCVPR2015} are out of the scope in this paper and comprehensively reviewed in~\cite{GengTPAMI2020}.

\textbf{Deep Learning Uncertainty.} To distinguish between the unknown and the known samples, an appropriate OOD scoring function is important. A recent line of research works~\cite{KrishnanNIPS2018,MalininNIPS2018,CharpentierNIPS2020,ShiNIPS2020,SensoyAAAI2020} show that the predictive uncertainty learned by deep neural networks (DNN) can be a promising scoring function to identify OOD samples. It is assumed that OOD samples should be highly uncertain during inference. Bayesian neural networks (BNN) has been introduced to model the epistemic and aleatoric uncertainty for multiple computer vision tasks~\cite{KendallNIPS2017,KrausITSC2019,BaoMM2020}. However, BNN is limited by the intractability of exact posterior inference, the difficulty of choosing suitable weight priors, and the expensive sampling for uncertainty quantification~\cite{AminiNIPS2020}. Recently, evidential deep learning (EDL) is developed by incorporating the evidential theory into deep neural networks with promising results in both classification~\cite{SensoyNIPS2018} and regression~\cite{AminiNIPS2020} tasks. In this paper, to the best of our knowledge, we are the first to incorporate evidential learning for large-scale and uncertainty-aware action recognition.

\textbf{Video Action Recognition.} Video action recognition has been widely studied in closed set setting~\cite{WuIJCNN2017,KongArXiv2018,zhangSensors2019}. In this paper, we select several representative and powerful methods, including the 3D convolution method I3D~\cite{I3DCVPR2017}, the 2D convolution method TSM~\cite{TSMICCV2019}, the two-stream method SlowFast~\cite{SlowFastICCV2019}, and the method focusing on neck structure of a recognition model TPN~\cite{TPNCVPR2020}. Note that our method can be easily applied to any existing video action recognition models to enable them for open set action recognition. 

\begin{figure}
    \centering
    \includegraphics[width=\linewidth]{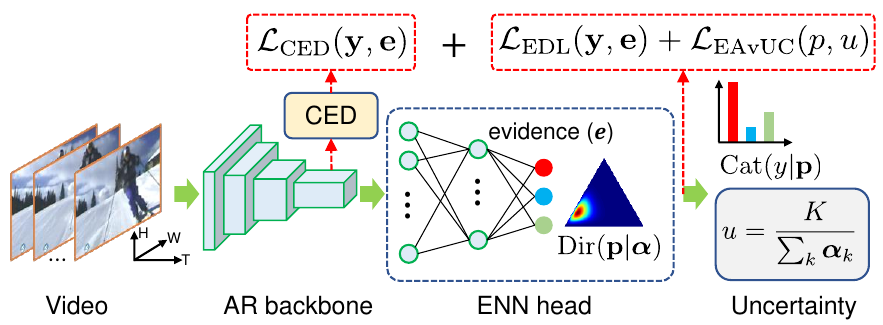}
    \captionsetup{font=small,aboveskip=1pt}
    \caption{The proposed DEAR method. We use 3-class ($K\!=\!3$) action recognition (AR) for illustration. On top of the AR backbone, the Evidential Neural Network (ENN) head predicts the evidence $\mathbf{e}$ to build the Dirichlet distribution of class probability $\mathbf{p}$. The evidential uncertainty ($u$) from the Dirichlet is used for rejecting the unknown in open set testing. 
    }
    \label{fig:framework}
\end{figure}

\section{Approach}

\textbf{Overview.} The proposed DEAR method is illustrated in Fig.~\ref{fig:framework}. Given a video as input, the Evidential Neural Network (ENN) head on top of an Action Recognition (AR) backbone\footnote{In our experiments, we use four different action recognition models which are I3D~\cite{I3DCVPR2017}, TSM~\cite{TSMICCV2019}, SlowFast~\cite{SlowFastICCV2019}, and TPN~\cite{TPNCVPR2020}.} predicts the class-wise evidence, which formulates a Dirichlet distribution so that the multi-class probabilities and predictive uncertainty of the input can be determined. For the open set inference, high uncertainty videos can be regarded as unknown actions while low uncertainty videos are classified by the learned categorical probabilities. The model is trained by Evidential Deep Learning (EDL)~\cite{SensoyNIPS2018} loss regularized by our proposed Evidential Uncertainty Calibration (EUC) method. In training, we also propose a plug-and-play Contrastive Evidence Debiasing (CED) module to debias the representation of human actions in videos. 

\subsection{Deep Evidential Action Recognition}

\textbf{Background of Evidential Deep Learning.} Existing deep learning-based models 
typically use a softmax layer on top of deep neural networks (DNNs) for classification. However, these softmax-based DNNs are not able to estimate the predictive uncertainty for a classification problem because softmax score is essentially a point estimation of a predictive distribution~\cite{YarinThesis2014} and the softmax outputs tend to be over-confident in false prediction~\cite{GuoICML2017}. 

Recent evidential deep learning (EDL)~\cite{SensoyNIPS2018,AminiNIPS2020} was developed to overcome the limitations of softmax-based DNNs by  introducing the evidence framework of Dempster-Shafer Theory (DST)~\cite{SentzBook2002} and the subjective logic (SL)~\cite{JosangBook2016}. EDL provides a principled way to jointly formulate the multi-class classification and uncertainty modeling. In particular, given a sample $\mathbf{x}^{(i)}$ for $K$-class classification, assuming that class probability follows a prior Dirichlet distribution, the cross-entropy loss to be minimized for learning evidence $\mathbf{e}^{(i)} \in \mathbb{R}_+^{K}$ eventually reduces to the following form:
\begin{equation}
    \mathcal{L}_{EDL}^{(i)}(\mathbf{y}^{(i)}, \mathbf{e}^{(i)}; \theta) = \sum_{k=1}^K \mathbf{y}_k^{(i)} \left(\log S^{(i)} - \log (\mathbf{e}_k^{(i)} + 1) \right)
\label{edlloss}
\end{equation}
where $\mathbf{y}^{(i)}$ is an one-hot $K$-dimensional label for sample $\mathbf{x}^{(i)}$ and $\mathbf{e}^{(i)}$ can be expressed as $\mathbf{e}^{(i)}\!=\!g\left(f(\mathbf{x}^{(i)};\theta)\right)$. Here, $f$ is the output of a DNN parameterized by $\theta$ and $g$ is the evidence function to keep evidence $\mathbf{e}_k$ non-negative. $S$ is the total strength of a Dirichlet distribution $\text{Dir}(\mathbf{p}|\boldsymbol{\alpha})$, which is parameterized by $\boldsymbol{\alpha}\in\mathbb{R}^K$, and $S$ is defined as $S\!=\!\sum_{k=1}^{K}\boldsymbol{\alpha}_k$. Based on DST and SL theory, the $\boldsymbol{\alpha}_k$ is linked to the learned evidence $\mathbf{e}_k$ by the equality $\boldsymbol{\alpha}_k=\mathbf{e}_k+1$. 
In the inference, the predicted probability of the $k$-th class is $\mathbf{\hat{p}}_k\!=\!\boldsymbol{\alpha}_k / S$ and the predictive uncertainty $u$ can be deterministically given as $u=K/S$. 
More detailed derivations could be found in our supplementary.

\textbf{EDL for Action Recognition.} In this paper, we propose to formulate the action recognition from the EDL perspective. In the training phase, by applying the EDL objective in~\eqref{edlloss} for action dataset, we are essentially trying to collect evidence of each action category for an action video. In the testing phase, since the action probability $\mathbf{p}\!\in\!\mathbb{R}^K$ is assumed to follow a Dirichlet, i.e., $\mathbf{p}\sim \text{Dir}(\mathbf{p}|\boldsymbol{\alpha})$, the categorical probability and uncertainty of a human action can be jointly expressed by a $(K-1)$-simplex (see the triangular heat map in Fig.~\ref{fig:framework}). The EDL uncertainty enables the action recognition model to ``know unknown".



However, due to the deterministic nature of EDL, the potential over-fitting issue would hamper the generalization capability for achieving good OSAR performance. Besides, the static bias problem in video data is still not addressed by EDL. To this end, we propose a model calibration method and a representation debiasing module below.



\subsection{Evidential Uncertainty Calibration}

\begin{figure}
    \centering
    \subcaptionbox{AC\label{simplex:ac}}{
    \includegraphics[width=0.21\linewidth]{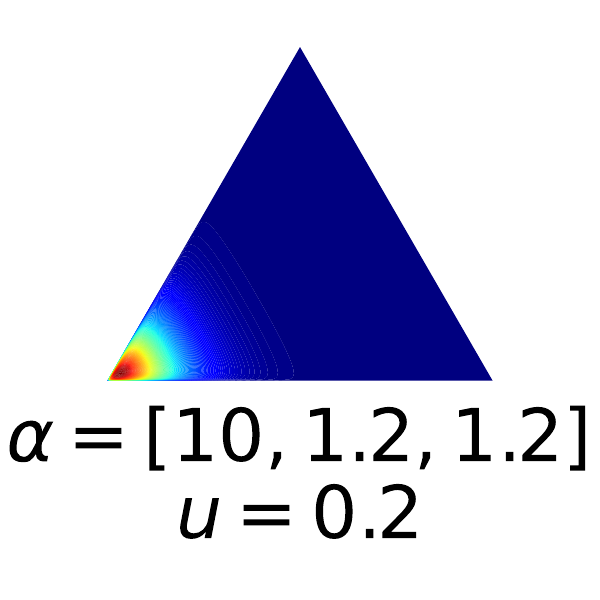}
    }
    \subcaptionbox{AU\label{simplex:au}}{
    \includegraphics[width=0.21\linewidth]{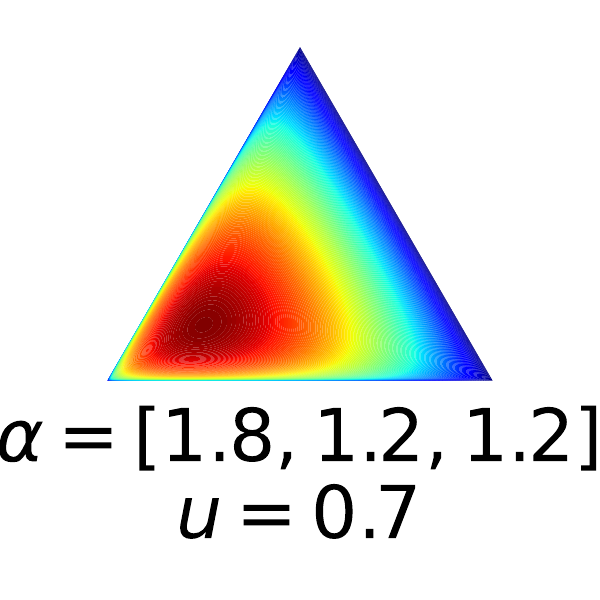}
    }
    \subcaptionbox{IC\label{simplex:ic}}{
    \includegraphics[width=0.21\linewidth]{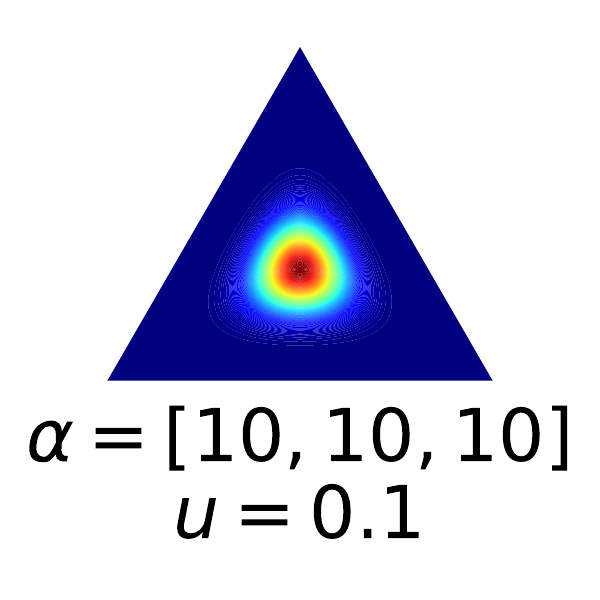}
    }
    \subcaptionbox{IU\label{simplex:iu}}{
    \includegraphics[width=0.21\linewidth]{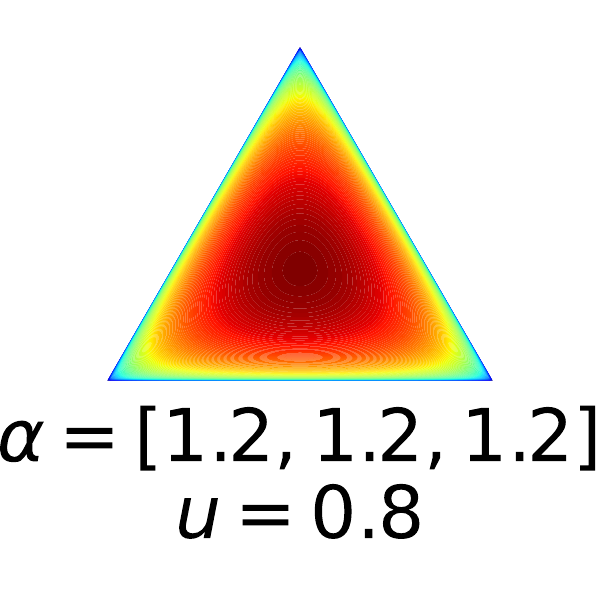}
    }
    \captionsetup{font=small,aboveskip=0pt}
    \caption{\textbf{Examples of Probability Simplex}. We use 3-class classification as an example and assume the first class as the correct label. A well calibrated model should give \textbf{A}ccurate and \textbf{C}ertain (AC) predictions (Fig.~\ref{simplex:ac}) or \textbf{I}naccurate and \textbf{U}ncertain (IU) predictions (Fig.~\ref{simplex:iu}), while the AU (Fig.~\ref{simplex:au}) and IC (Fig.~\ref{simplex:ic}) cases need to be reduced.}
    \label{fig:simplex}
\end{figure}
Though the evidential uncertainty from EDL can be directly learned without sampling, the uncertainty may not be well calibrated to handle the unknown samples in OSAR setting. As pointed out in existing model calibration literature~\cite{MukhotiArXiv2018,KrishnanNIPS2020}, a well calibrated model should be confident in its predictions when being accurate, and be uncertain about inaccurate ones. Besides, existing DNN models have been empirically demonstrated that miscalibration is linked to the over-fitting of the negative log-likelihood (NLL)~\cite{GuoICML2017,MukhotiNIPS2020}. Since the EDL objective in \eqref{edlloss} is equivalent to minimizing the NLL~\cite{SensoyNIPS2018}, the trained model is likely to be over-fitted with poor generalization for OSAR tasks. To address this issue, we propose to calibrate the EDL model by considering the relationship between the accuracy and uncertainty.

To this end, we follow the same goal as~\cite{MukhotiArXiv2018,KrishnanNIPS2020} to maximize the \textit{Accuracy versus Uncertainty} (AvU) utility function for calibrating the uncertainty:
\begin{equation}
    \text{AvU} = \frac{n_{AC} + n_{IU}}{n_{AC} + n_{AU} + n_{IC} + n_{IU}}
\label{avu}
\end{equation}
where the $n_{AC}$, $n_{AU}$, $n_{IC}$, and $n_{IU}$ represent the numbers of samples in four predicted cases, i.e., (1) \textbf{A}ccurate and \textbf{C}ertain (\textbf{AC}), (2) \textbf{A}ccurate and \textbf{U}ncertain (\textbf{AU}), (3) \textbf{I}naccurate and \textbf{C}ertain (\textbf{IC}), and (4) \textbf{I}naccurate and \textbf{U}ncertain (\textbf{IU}). A well calibrated model could achieve high AvU utility so that the predictive uncertainty can be consistent with accuracy. Fig.~\ref{fig:simplex} shows a toy example of the four possible EDL outputs. To calibrate the predictive uncertainty, the EDL model is encouraged to learn a skewed and sharp Dirichlet simplex for accurate prediction (Fig.~\ref{simplex:ac}), and to provide an unskewed and flat Dirichlet simplex for incorrect prediction (Fig.~\ref{simplex:iu}). To this end, we propose to regularize EDL training by minimize the expectations of AU and IC cases (Fig.~\ref{simplex:au} and Fig.~\ref{simplex:ic}) such that the other two cases can be encouraged. Therefore, if a video is assigned with high EDL uncertainty, it is more likely to be incorrect so that an unknown action is identified. 

In particular, we propose an \textit{Evidential Uncertainty Calibration} (EUC) method to minimize the following sum of AU and IC cases by considering the logarithm constraint between the confidence $p_i$ and uncertainty $u_i$: 
\begin{equation}
\begin{split}
    \mathcal{L}_{EUC} = -\lambda_t & \sum_{i\in \{\hat{y}_i = y_i\}} p_i \log (1-u_i)  \\
    - (1-\lambda_t) & \sum_{i\in \{\hat{y}_i \neq y_i\}} (1-p_i)\log (u_i)
\end{split}
\label{eq:loss_euc}
\end{equation}
where $p_i$ is the maximum class probability of an input sample $\mathbf{x}^{(i)}$ and $u_i$ is the associated evidential uncertainty. The first term aims to give low uncertainty ($u_i\!\rightarrow\! 0$) when the model makes accurate prediction ($\hat{y}_i\!=\!y_i, p_i\rightarrow 1$), while the second term tries to give high uncertainty ($u_i \!\rightarrow\! 1$) when the model makes inaccurate prediction ($\hat{y}_i\!\neq\! y_i, p_i\rightarrow 0$). 
Note that the annealing factor $\lambda_t\in [\lambda_0, 1]$ is defined as $\lambda_t \!=\! \lambda_0 \exp \left\{-(\ln{\lambda_0}/T) t\right\}$. Here, $\lambda_0$ is a small positive constant, i.e., $\lambda_0\! \ll\! 1$, such that $\lambda_t$ is monotonically increasing w.r.t. training epoch $t$, and $T$ is the total number of training epochs. As the training epoch $t$ increasing to $T$, the factor $\lambda_t$ will be exponentially increasing from $\lambda_0$ to 1.0. 

The motivation behind the annealing weighting is that the dominant periods of accurate and inaccurate predictions in model training are different. In the early training stages, the inaccurate predictions are the dominant cases so that the IC loss (second term) should be more penalized, while in the late training stages, the accurate predictions are the dominant so that the AU loss (first term) should be more penalized. Therefore, the annealing weighing factor $\lambda_t$ dynamically balances the two terms in training. 


\textbf{Discussion.} Our EUC method is advantageous over existing approaches~\cite{MukhotiArXiv2018} and AvUC~\cite{KrishnanNIPS2020} in following aspects. First, compared with~\cite{MukhotiArXiv2018}, our EUC method takes the same merit of AvUC that it is a fully differentiable regularization term. Second, compared with both~\cite{MukhotiArXiv2018} and AvUC, the EUC loss does not rely on distribution shifted validation set during training which is not reasonable for OSAR model to access the OOD samples. Therefore, our method provides better flexibility to calibrate deep learning models on large-scale dataset, such as the real-world videos of human actions addressed in this paper. 
Our experimental results (Table~\ref{tab:ece}) show that the model calibration performance of EUC method is more significant for open set recognition than on closed set recognition.


\subsection{Contrastive Evidence Debiasing}

For OSAR task, static bias (see example in Fig.~\ref{fig:bias}) in a video dataset is one of the most challenging problems that limit the generalization capability of a model in an open set setting. According to~\cite{LiECCV2018}, static bias can be categorized into scene bias, object bias, and human bias. Existing research work~\cite{ChoiNIPS2019,LiECCV2018,KimCVPR2019,BahngICML2020} has empirically shown that debiasing the model by input data or learned representation can significantly improve the action recognition performance. As pointed out in~\cite{LiECCV2018}, it is intrinsically nothing wrong about the bias if it can be ``over-fitted" by an action recognition model for achieving a ``good" performance in traditional closed-set setting. However, in an open set setting, the static bias could result in a vulnerable model that falsely recognizes an action video containing similar static features but totally out-of-contextual temporal dynamics.

\begin{figure}
    \centering
    \includegraphics[width=\linewidth]{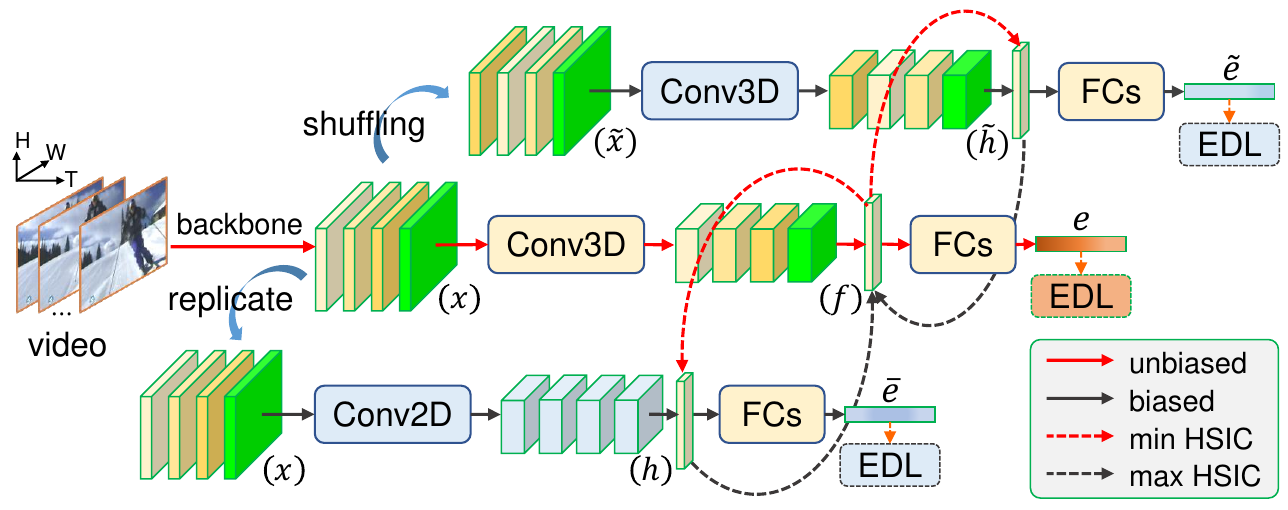}
    \captionsetup{font=small,aboveskip=0pt}
    \caption{\textbf{Contrastive Evidence Debiasing (CED) Module}. The module consists of three branches with similar structure. In contrast to the middle branch, the top and bottom ones aim to learn a biased evidence by temporally shuffled feature input and 2D convolution (\textit{Conv2D}), respectively. The generated feature $\mathbf{f}$ is contrastively pushed to be independent of biased feature $\mathbf{h}$.}
    \label{fig:ced}
\end{figure}

In this paper, we propose a Contrastive Evidence Debiasing (CED) module to mitigate the static bias problem. As shown in Fig.~\ref{fig:ced}, the CED consists of three branches. The middle branch is a commonly-used 3D convolutional structure (Conv3D) to predict unbiased evidence ($\mathbf{e}$) while the top and and bottom branches predict biased evidences ($\mathbf{\Tilde{e}}$ and $\mathbf{\bar{e}}$). In particular, the top branch keeps the same network structure as the middle one but takes temporally shuffled features ($\mathbf{\Tilde{x}}$) as input. The bottom branch keeps the same input feature ($\mathbf{x}$) as the middle one but replaces the Conv3D with 2D convolutional structure (Conv2D). Finally, with the HSIC-based minmax optimization, the feature $\mathbf{f}$ for predicting unbiased evidence is encouraged to be contrastive to the features $\mathbf{h}$ and $\mathbf{\Tilde{h}}$ for predicting biased evidence. 

In particular, motivated by the recent method ReBias~\cite{BahngICML2020}, the minmax optimization is defined by using the Hilbert-Schmidt Independence Criterion (HSIC). The HSIC function measures the degree of independence between two continuous random variables. With radial basis function (RBF) kernel $k_1$ and $k_2$, $\text{HSIC}^{k_1,k_2}(\mathbf{f}, \mathbf{h})\!=\!0$ if and only if $\mathbf{f}\indep \mathbf{h}$. The detailed mathematical form of HSIC can be found in~\cite{GrettonICALT2005,SongJMLR2012} (or see the Section 1.3 of the supplementary). For the middle branch, the goal is to learn a discriminative and unbiased feature $\mathbf{f}$ by minimizing
\begin{equation}
    \mathcal{L}(\theta_f, \phi_f) = \mathcal{L}_{EDL}(\mathbf{y},\mathbf{e};\theta_{f}, \phi_f) + \lambda \sum_{\mathbf{h}\in \Omega} \text{HSIC}(\mathbf{f}, \mathbf{h}; \theta_f),
\label{debias}
\end{equation}
where $\theta_f$ and $\phi_f$ are parameters of neural networks to produce unbiased feature $\mathbf{f}$ and to predict evidence $\mathbf{e}$. $\mathbf{y}$ is the multi-class label. The second term encourages feature $\mathbf{f}$ to be independent of the biased feature $\mathbf{h}$ from the set of features generated by top branch $h_{3D}(\mathbf{\Tilde{x}})$ and the bottom branch $h_{2D}(\mathbf{x})$, i.e., $\Omega\!=\!\{h_{3D}(\mathbf{\Tilde{x}}), h_{2D}(\mathbf{x})\}$.

For the top and bottom branches, the goal is to learn the above two types of biased feature $\mathbf{h}$ by
\begin{equation}
    \mathcal{L}(\theta_h, \phi_h) = \sum_{\mathbf{h}\in \Omega} \left\{ \mathcal{L}_{EDL}(\mathbf{y},\mathbf{e}_h;\theta_{h}, \phi_h) - \lambda \text{HSIC}(\mathbf{f}, \mathbf{h}; \theta_h)\right\}
\label{bias}
\end{equation}
where $\theta_h$ denotes the network parameters of $h_{3D}(\mathbf{\Tilde{x}})$ and $h_{2D}(\mathbf{x})$ to generate biased features $\mathbf{h}$, and the $\phi_h$ denotes the parameters of neural networks to predict corresponding evidence $\mathbf{e}_h \in \{\hat{\mathbf{e}},\bar{\mathbf{e}}\}$. The first term in~\eqref{bias} aims to avoid the biased feature $\mathbf{h}$ to predict arbitrary evidence, while the second term guarantees that $\mathbf{h}$ is similar enough to $\mathbf{f}$ so that $\mathbf{f}$ has to be pushed faraway from $\mathbf{h}$ by~\eqref{debias}.

The two objectives in \eqref{debias} and \eqref{bias} are alternatively optimized so that feature $\mathbf{h}$ is learned to be biased to guide the debiasing of feature $\mathbf{f}$. In practice, we also implemented a joint training strategy which aims to optimize the objective of \eqref{debias} and \eqref{bias} jointly and we empirically found it can achieve a better performance.

\textbf{Discussion.} Compared with recent work~\cite{ChoiNIPS2019} that leverages adversarial learning to remove scene bias, our method does not rely on object bounding boxes and pseudo scene labels as auxiliary training input. The representation bias addressed in our paper implicitly encompasses all sources of biases, not just the scene bias. 
Compared with ReBias~\cite{BahngICML2020}, our CED module shares the similar idea of removing bias with bias. However, the HSIC in our CED module considers not only the bias-characterising model (i.e., $h_{2D}(\mathbf{x})$) as in~\cite{BahngICML2020}, but also the biased feature input by temporal shuffling. This consideration will further encourage the backbone to focus more on temporal dynamics. Besides, 
our CED is a plug-and-play module and can be flexibly inserted into any state-of-the-art deep learning-based action recognition models with little coding effort.

\section{Experiments}

\begin{table*}[!htp]
\centering
\captionsetup{font=small,aboveskip=3pt}
\caption{\textbf{Comparison with state-of-the-art methods.} Models are trained on the closed set UCF-101~\cite{ucf101} and tested on two different open sets where the samples of unknown class are from HMDB-51~\cite{hmdb51} and MiT-v2~\cite{mitv2}, respectively. For Open maF1 scores, both the mean and standard deviation of 10 random trials of unknown class selection are reported. Closed set accuracy is for reference only.}
\label{tab:sota}
\small
\setlength{\tabcolsep}{1.0mm}
\setlength{\extrarowheight}{0.5mm}
\begin{tabular}{l|l|cc|cc|c}
\hline
\multirow{2}{*}{Models} &\multirow{2}{*}{OSAR Methods}  &\multicolumn{2}{c|}{UCF-101~\cite{ucf101} + HMDB-51~\cite{hmdb51}} &\multicolumn{2}{c|}{UCF-101~\cite{ucf101} + MiT-v2~\cite{mitv2}} &\multirow{2}{*}{\makecell[c]{Closed Set Accuracy (\%) \\ (\textbf{For reference only})}} \\
\cline{3-6}
& &Open maF1 (\%) & Open Set AUC (\%) &Open maF1 (\%) &Open Set AUC (\%) & \\
\hline
\multirow{6}{*}{I3D~\cite{I3DCVPR2017}} 
&OpenMax~\cite{BendaleCVPR2016} & 67.85 $\pm$ 0.12 & 74.34 & 66.22 $\pm$ 0.16 & 77.76 & 56.60 \\
&MC Dropout  &71.13 $\pm$ 0.15 & 75.07&68.11 $\pm$ 0.20 &79.14 & 94.11\\
&BNN SVI~\cite{KrishnanNIPS2018}  &71.57 $\pm$ 0.17 &74.66 &68.65 $\pm$ 0.21 &79.50 &93.89\\
&SoftMax   & 73.19 $\pm$ 0.17 & 75.68 & 68.84 $\pm$ 0.23 & 79.94 & 94.11\\
&RPL~\cite{ChenECCV2020}  & 71.48 $\pm$ 0.15 & 75.20 & 68.11 $\pm$ 0.20 & 79.16 & 94.26 \\
&DEAR (ours)  &\textbf{77.24 $\pm$} 0.18 &\textbf{77.08} &\textbf{69.98} $\pm$ 0.23 &\textbf{81.54} & 93.89\\
\hline
\multirow{6}{*}{TSM~\cite{TSMICCV2019}} 
&OpenMax~\cite{BendaleCVPR2016}  & 74.17 $\pm$ 0.17 & 77.07& \textbf{71.81} $\pm$ 0.20 & 83.05 & 65.48\\
&MC Dropout  &71.52 $\pm$ 0.18 & 73.85&65.32  $\pm$ 0.25 & 78.35 &95.06\\
&BNN SVI~\cite{KrishnanNIPS2018}  & 69.11 $\pm$ 0.16 & 73.42 & 64.28 $\pm$ 0.23 & 77.39 & 94.71 \\
&SoftMax   & 78.27 $\pm$ 0.20 & 77.99 & 71.68 $\pm$ 0.27 & 82.38 & 95.03\\
&RPL~\cite{ChenECCV2020}  & 69.34 $\pm$ 0.17  & 73.62 & 63.92 $\pm$ 0.25  & 77.28 & 95.59 \\
&DEAR (ours)  &\textbf{84.69} $\pm$ 0.20 &\textbf{78.65} &70.15 $\pm$ 0.30 &\textbf{83.92} &94.48\\
\hline
\multirow{6}{*}{SlowFast~\cite{SlowFastICCV2019}} 
&OpenMax~\cite{BendaleCVPR2016}  & 73.57 $\pm$ 0.10 & 78.76 & 72.48 $\pm$ 0.12 & 80.62 & 62.09\\
&MC Dropout  &70.55 $\pm$ 0.14 &75.41 &67.53 $\pm$ 0.17 & 78.49 &96.75\\
&BNN SVI~\cite{KrishnanNIPS2018}  & 69.19 $\pm$ 0.13 & 74.78 & 65.22 $\pm$ 0.21 & 77.39 & 96.43 \\
&SoftMax   & 78.04 $\pm$ 0.16 & 79.16 & 74.42 $\pm$ 0.22 & 82.88 & 96.70\\
&RPL~\cite{ChenECCV2020}  & 68.32 $\pm$ 0.13 & 74.23 & 66.33 $\pm$ 0.17 & 77.42 & 96.93 \\
&DEAR (ours)  &\textbf{85.48} $\pm$ 0.19 &\textbf{82.94} &\textbf{77.28} $\pm$ 0.26 &\textbf{86.99} &96.48\\
\hline
\multirow{6}{*}{TPN~\cite{TPNCVPR2020}} 
&OpenMax~\cite{BendaleCVPR2016} & 65.27 $\pm$ 0.09 & 74.12 & 64.80 $\pm$ 0.10 & 76.26 & 53.24\\
&MC Dropout  &68.45 $\pm$ 0.12 &74.13 &65.77 $\pm$ 0.17 &77.76 &95.43\\
&BNN SVI~\cite{KrishnanNIPS2018} &63.81 $\pm$ 0.11 &72.68 &61.40 $\pm$ 0.15 &75.32 & 94.61\\
&SoftMax   & 76.23 $\pm$ 0.14 & 77.97 & 70.82 $\pm$ 0.21 & 81.35 & 95.51\\
&RPL~\cite{ChenECCV2020}  & 70.31 $\pm$ 0.13 & 75.32 & 66.21 $\pm$ 0.21 & 78.21 & 95.48 \\
&DEAR (ours)  &\textbf{81.79} $\pm$ 0.15 &\textbf{79.23} &\textbf{71.18} $\pm$ 0.23 &\textbf{81.80} &96.30\\
\hline
\end{tabular}
\end{table*}

\textbf{Dataset.} We evaluate the proposed DEAR method on three commonly used real-world video action datasets, including UCF-101~\cite{ucf101}, HMDB-51~\cite{hmdb51}, and MiT-v2~\cite{mitv2}. All models are trained on UCF-101 training split. MiT-v2 has 305 classes and its testing split contains 30,500 video samples, which are about 20 times larger than the HMDB-51 testing set. In testing, we use the UCF-101 testing set as known samples, and the testing splits of HMDB-51 and MiT-v2 datasets as two sources of unknown. Note that there could be a few overlapping classes between UCF-101 and the other two datasets, but for standardizing the evaluation and reproducibility, we do not manually clean the data. 

\textbf{Evaluation Protocol.} To evaluate the classification performance on both closed and open set settings, we separately report the Closed Set Accuracy for $K$-class classification and the Open Set area under ROC curve (AUC) for distinguishing known and unknown (2 classes). Furthermore, to comprehensively evaluate the $(K+1)$-class classification performance, i.e., the unknown as the $(K+1)$-th class, we plot the curve of macro-F1 scores by gradually increasing the openness similar to existing literature~\cite{ShuICME2018,yoshihashiCVPR2019,SunCVPR2020}. For each openness point, $i$ new classes are randomly selected from HMDB-51 (where $i\!\leq\!51$) or MiT-v2 (where $i\!\leq\!305$) test set and we compute the macro-F1 score for each of 10 randomized selections. Since there is no existing quantitative metric to summarize the performance of the F1 curve, in this paper we propose an \textbf{Open maF1} score:
\begin{equation}
    \text{Open maF1} = \frac{\sum_i \omega_O^{(i)}\cdot F_1^{(i)}}{\sum_i \omega_O^{(i)}}
\end{equation}
where $\omega_O^{(i)}$ denotes the openness when $i$ new classes are introduced and it is defined as $\omega_O^{(i)}\!=\!1\!-\!\sqrt{2K / (2K+i)}$ according to~\cite{ScheirerTPAMI2012}. $F_1^{(i)}$ is the macro-F1 score by considering the samples from all new classes as unknown. The basic idea of weighting $F_1$ by $\omega_O$ is that the result is essentially the normalized area under the curve of macro-F1 vs. openness. The Open maF1 quantitatively evaluates the performance of $(K+1)$-class classification in open set setting.


\textbf{Implementation Details.} Our method is implemented with the PyTorch codebase MMAction2~\cite{2020mmaction2}. 
The adopted action models are experimented with ResNet-50 backbone pre-trained on Kinetics-400~\cite{kinetics} dataset and fine-tuned on UCF-101 training set. Our proposed EDL loss $\mathcal{L}_{EDL}$ is used to replace the original cross-entropy loss, and our proposed CED module is inserted into the layer before the classification heads of recognition models. During training, we use base learning rate 0.001 and it is step-wisely decayed for every 20 epochs with totally 50 epochs. We set batch size as 8 during training. The rest of hyperparameters are kept the same as the default configuration provided by MMAction2. During inference, our CED module is removed. Other implementation details are provided in the supplementary.

\begin{figure}[t]
    \centering
    \subcaptionbox{HMDB-51 as Unknown}{
    \includegraphics[width=0.47\linewidth]{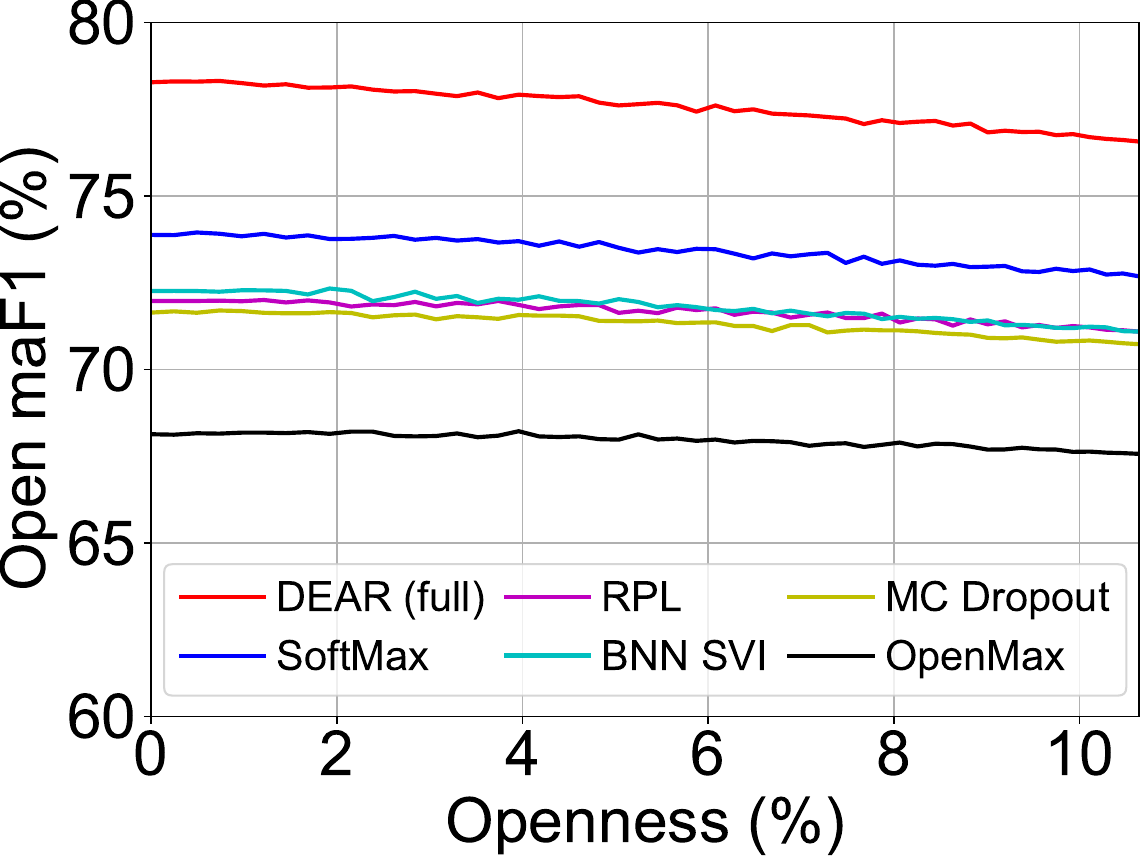}
    }
    \subcaptionbox{MiT-v2 as Unknown}{
    \includegraphics[width=0.47\linewidth]{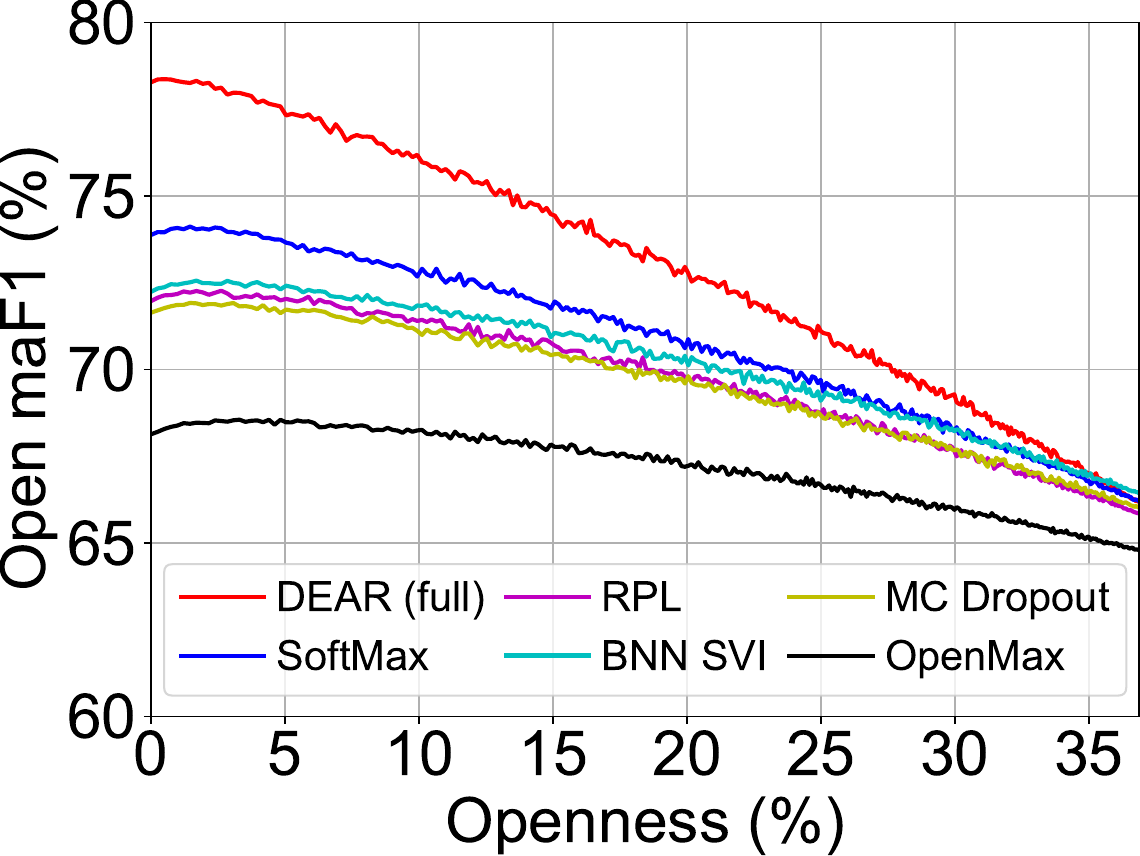}
    }
    \captionsetup{font=small,aboveskip=0pt}
    \caption{\textbf{Open macro-F1 scores against varying Openness}. The maximum openness is determined by the number of unknown classes, i.e., in $\omega_O^{(i)}$, $i\!=\!51$ for HMDB-51 and $i\!=\!305$ for MiT-v2.}
    \label{fig:f1_openness}
\end{figure}

\subsection{Comparison with State-of-the-art}

The proposed DEAR method is compared with baselines as shown in the second column of Table~\ref{tab:sota}. The open set performances are also summarized in Fig.~\ref{fig:gain_curve}. For these baselines, SoftMax, OpenMax, and MC Dropout share the same trained model since they are only different in testing phase. For the MC Dropout and BNN SVI which incorporate stochastic sampling in testing, we set the 10 forward passes through the model and adopt the BALD~\cite{HoulsbyArXiv2011} method to quantify the model uncertainty as suggested by~\cite{KrishnanNIPS2018}. 
Following~\cite{SunCVPR2020}, the threshold of scoring function is determined by ensuring 95\% training data to be recognized as known.

\textbf{Open Set Action Recognition.} In Table~\ref{tab:sota}, we report the results of both closed set and open set performance. 
It shows that with different action recognition models, our method consistently and significantly outperforms baselines on Open maF1 score for $(K\!+\!1)$-class classification and Open Set AUC score for rejecting the unknowns, while only sacrifices less than 1\%  performance decrease on Closed Set Accuracy. When equipped with SlowFast model, our method could improve the MC Dropout method almost 8\% of open set AUC and 15\% of Open maF1 score. OpenMax and RPL are the recent state-of-the-art OSR methods, however we find that their performances are far behind our DEAR method on the OSAR task. Note that the closed set accuracy of OpenMax is dramatically lower than other baselines, this is because OpenMax directly modifies the activation layer before softmax and appends the unknown class as output, which could destroy the accurate predictions of known samples. Besides, we also note that with TSM model, the Open maF1 score of DEAR method is slightly inferior to OpenMax on MiT-v2 dataset. This indicates that for large-scale unknown testing data such as MiT-v2, the 2D convolution-based TSM is not a good choice for the DEAR method as compared to those 3D convolution-based architectures such as I3D, SlowFast, and TPN.  

Based on I3D model, as depicted in Fig.~\ref{fig:f1_openness}, we plot the average Open maF1 scores against varying openness by incrementally introducing HMDB-51 and MiT-v2 testing sets as unknown. It clearly shows that the proposed DEAR method achieves the best performance. Note that for the large scale MiT-v2 dataset, as the openness increasing, the performances of different methods converge to be closed to each other. This is because the macro-F1 is sensitive to class imbalance and it will be gradually dominated by the increasing unknown classes from totally 305 categories in MiT-v2. Nevertheless, our method DEAR still keeps better than all other baselines.

\begin{figure}[t]
    \centering
    \subcaptionbox{MC Dropout}{
        \includegraphics[width=0.465\linewidth]{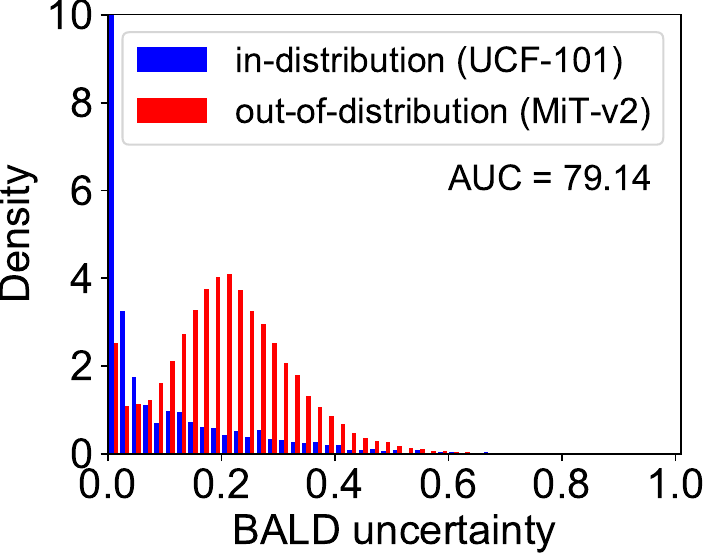}
    }
    \subcaptionbox{BNN SVI~\cite{KrishnanNIPS2018}}{
        \includegraphics[width=0.465\linewidth]{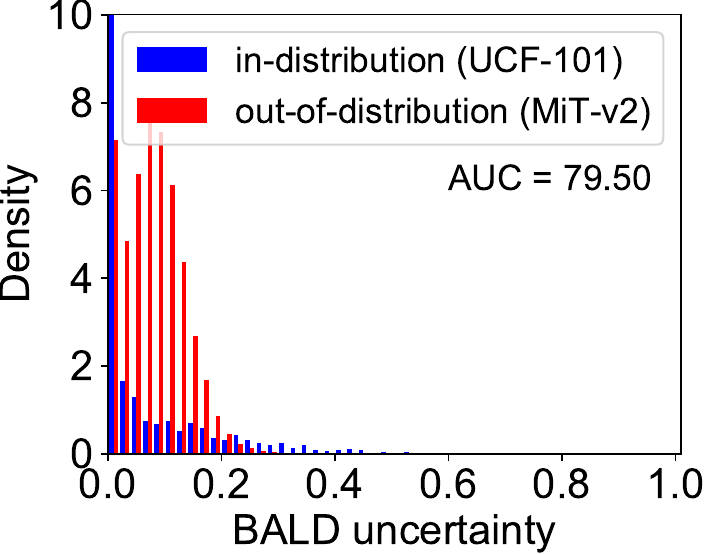}
    }
    \subcaptionbox{DEAR (vanilla)}{
        \includegraphics[width=0.465\linewidth]{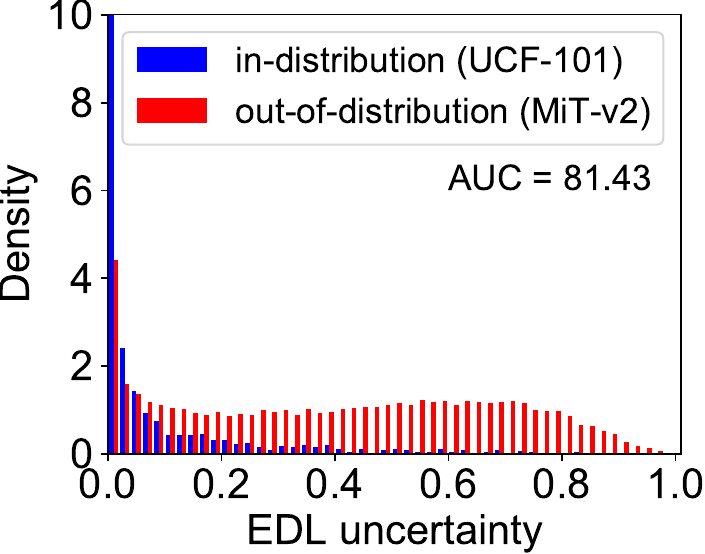}
    }
    \subcaptionbox{DEAR (full)}{
        \includegraphics[width=0.465\linewidth]{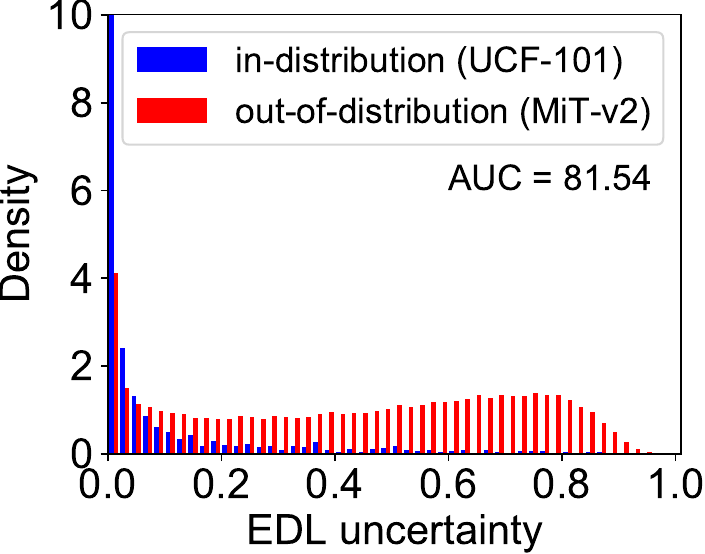}
    }
    \captionsetup{font=small,aboveskip=0pt}
    \caption{\textbf{Out-of-distribution Detection by Uncertainty.} The DEAR (vanilla) is the variant of DEAR (full) that only $\mathcal{L}_{EDL}$ is used for model training. We use MiT-v2 as unknown and I3D as the recognition model. Uncertainty values are normalized to [0,1] within each distribution.}
    \label{fig:ood_mit}
\end{figure}

\textbf{Out-of-distribution Detection. } 
This task aims to distinguish between the in-distribution samples (known) and out-of-distribution (OOD) samples (unknown). Similar to the baseline MC Dropout and BNN SVI~\cite{KrishnanNIPS2018}, which are using uncertainty as scoring function to identify the unknown, the OOD detection performance can be evaluated by showing the Open Set AUC in Table~\ref{tab:sota} and the histogram statistics in Fig.~\ref{fig:ood_mit}. The AUC numbers and figures clearly show that our DEAR method with EDL uncertainty can better detect the OOD samples. Compared with the vanilla DEAR which only uses $\mathcal{L}_{EDL}$ for model training, the estimated uncertainties of OOD samples skews closer to 1.0. 
More results can be found in our supplementary materials.

\subsection{Ablation Study}
\label{sec:ablation}

\textbf{Contribution of Each Component.} In Table~\ref{tab:ablation}, it shows the OSAR performance of each DEAR variant. The experiments are conducted with TPN model and evaluated using HMDB-51 testing set as unknown. The results demonstrate that all the proposed components could contribute to the OSAR performance gain. In particular, the $h_{2D}(\mathbf{x})$ of our CED module contributes the most. Besides, the joint training of CED module shows slightly better than the alternative training. Therefore, by default the joint training is adopted throughout other experiments.

\textbf{Model Calibration.} Though the proposed EUC module can improve the performance on OSAR task (as shown in Table~\ref{tab:ablation}), we further dig into the question that if the performance gain of EUC results from better calibrating a classification model. To this end, we adopt the widely used Expected Calibration Error (ECE)~\cite{GuoICML2017} to evaluate the model calibration performance of our full method DEAR (full) and its variant without EUC loss $\mathcal{L}_{EUC}$. Quantitative results are reported in Table~\ref{tab:ece}. It shows that $\mathcal{L}_{EUC}$ can reduce the ECE values with both open set and closed set recognition settings. In particular, the calibration capability is more significant in open set setting than in closed set setting. This validates our claim that the proposed $\mathcal{L}_{EUC}$ could calibrate an OSAR model. 

\begin{table}[t]
\centering
\setlength{\tabcolsep}{2.0mm}
\setlength{\extrarowheight}{1.5mm}
\captionsetup{font=small,aboveskip=3pt}
\caption{\textbf{Ablation studies.} Based on TPN~\cite{TPNCVPR2020} model, HMDB-51~\cite{hmdb51} is used as the unknown. Best results are shown in bold.}
\label{tab:ablation}
\footnotesize
\begin{tabular}{c|c|c|c|c}
\hline
$\mathcal{L}_{EUC}$ & CED & Joint Train & Open maF1 (\%) & OS-AUC (\%) \\
\hline
\xmark &\xmark  & \checkmark &74.95 $\pm$ 0.18 &77.12 \\
\checkmark &\xmark  & \checkmark &75.88 $\pm$ 0.16 &77.49 \\
\checkmark &\checkmark  & \xmark & 81.18 $\pm$ 0.15& 79.02 \\
\checkmark &\checkmark &\checkmark & \textbf{81.79} $\pm$ \textbf{0.15} &\textbf{79.23} \\
\hline
\end{tabular}
\end{table}

\begin{table}[t]
\centering
\setlength{\tabcolsep}{1mm}
\setlength{\extrarowheight}{0.5mm}
\captionsetup{font=small,aboveskip=3pt}
\caption{\textbf{Expected Calibration Error (ECE) results.} Small ECE indicates the model is better calibrated. The numbers in brackets indicate the number of classes involved in evaluation.}
\label{tab:ece}
\footnotesize
\begin{tabular}{l|c|c|c}
\hline
Model variants &Open Set (K+1) &Open Set (2) &Closed Set (K) \\
\hline
DEAR (w/o $\mathcal{L}_{EUC}$) &0.284 &0.256 &0.030 \\
DEAR (full) & \textbf{0.268} &\textbf{0.239} &\textbf{0.029} \\
\hline
\end{tabular}
\end{table}

\begin{table}[t]
\centering
\setlength{\tabcolsep}{2.0mm}
\setlength{\extrarowheight}{0.5mm}
\captionsetup{font=small,aboveskip=3pt}
\caption{\textbf{Accuracy (\%) on Biased and Unbiased dataset.}}
\label{tab:bias}
\small
\begin{tabular}{l|cc|cc}
\hline
\multirow{2}{*}{Methods} &\multicolumn{2}{c|}{Biased (Kinetics)} &\multicolumn{2}{c}{Unbiased (Mimetics)} \\
\cline{2-5}
&top-1 &top-5 &top-1 &top-5 \\
\hline
DEAR (w/o CED) &91.18 &99.30 &26.56 &69.53 \\
DEAR (full) &\textbf{91.18} &\textbf{99.54} &\textbf{34.38} &\textbf{75.00} \\
\hline
\end{tabular}
\end{table}

\textbf{Representation Debiasing.} To further validate if the performance gain of our CED module is rooted in the representation debiasing, we use Kinetics~\cite{kinetics} as a biased dataset and Mimetics~\cite{mimetics} as an unbiased dataset. Similar to~\cite{BahngICML2020}, we select 10 human action categories from Kinetics for training and biased testing, and select the same categories from Mimetics for unbiased testing. Without the pre-trained model from Kinetics dataset, we apply our DEAR method with and without CED on TSM model. The top-1 and top-5 accuracy results are reported in Table~\ref{tab:bias}. It shows that models trained on biased dataset (Kinetics) are vulnerable on unbiased dataset (Mimetics). However, when equipped with the proposed CED module, the performance on the unbiased dataset can be significantly improved while performance on the biased dataset still keeps minor changes.

\begin{figure}
    \centering
    \includegraphics[width=0.8\linewidth]{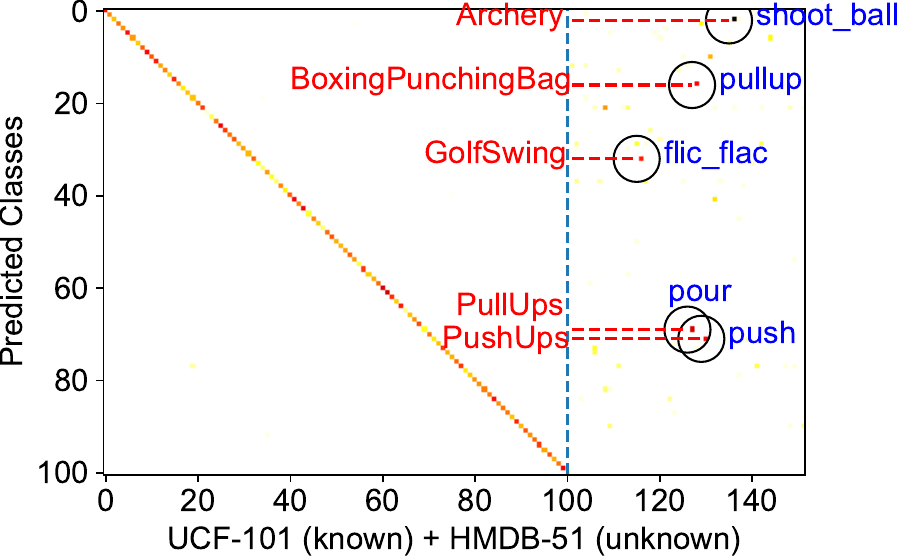}
    \captionsetup{font=small,aboveskip=5pt}
    \caption{\textbf{Confusion Matrix for Known and Unknown.} 
    The $x$-axis shows the ground truth classes of both UCF-101 (known) and HMD-51 (unknown), and $y$-axis represents the predicted classes defined by UCF-101. This figure highlights the top-5 unknown classes (blue text) that are mis-classified as the known (red text).}
    \label{fig:confmats}
\end{figure}

\textbf{What Types of Unknown are Mis-classified?} As shown in Fig.~\ref{fig:confmats}, the confusion matrix is visualized by considering both the known classes from UCF-101 and unknown classes from HMDB-51 datasets. It shows that in spite of high closed set accuracy (the diagonal line), the actions from unknown classes could be easily classified as known categories. For example, \textit{shoot ball} is the top-1 mis-classified unknown class in HMDB-51, which is the most frequently mis-classified as the known class \textit{Archery} in UCF-101. It is convincing that the mis-classification is caused by their similar background scene, i.e., large area of grass land, which is static bias as addressed in this paper.

\section{Conclusion}

In this paper, we proposed a Deep Evidential Action Recognition (DEAR) method for the open set action recognition (OSAR) problem. OSAR is more challenging than image OSR problem due to the uncertain nature of temporal action dynamics and the static bias of background scenes. 
To this end, we conduct  Evidential Deep Learning (EDL) to learn a discriminative action classifier with quantified predictive uncertainty, where the uncertainty is used to distinguish between the known and unknown samples. As novel extensions of EDL, an Evidential Uncertainty Calibration (EUC) method and a contrastive evidential debiasing (CED) module are proposed to address the unique challenges in OSAR. Extensive experimental results demonstrate that our DEAR method works for most existing action recognition models in open set setting. 

\textbf{Acknowledgement}. 
This research is supported by an ONR Award N00014-18-1-2875 and the Army Research Office under grant number W911NF-21-1-0236.
The views and conclusions contained in this document are those of the authors and should not be interpreted as representing the official policies, either expressed or implied, of the Office of Naval Research, the Army Research Office or the U.S. Government.
We especially thank Google Cloud Platform providing two NVIDIA A100 SXM4 GPUs.

{\small
\bibliographystyle{ieee}
\bibliography{ref}
}

\clearpage
\begin{appendix}

\section*{Supplementary Material}

In this document, additional materials are provided to supplement our main paper. In section~\ref{prelim}, the preliminary knowledge about the evidential deep learning and model calibration are described in detail, which are helpful to understand the methodology of our main paper. In section~\ref{implem}, additional implementation details are provided, which are useful to reproduce our proposed method. Sections~\ref{quanti} and~\ref{quali} provide additional experimental results to complement the ones presented in our main paper.

\section{Detailed Methodology}
\label{prelim}

\subsection{Preliminaries of Evidential Deep Learning}


Existing video action recognition models typically use softmax on top of deep neural networks (DNN) for classification. However, the softmax function is heavily limited in the following aspects. First, the predicted categorical probabilities have been squashed by the denominator of softmax. This is known to result in an over-confident prediction for the unknown data, which is even more detrimental to open set recognition problem than the closed set recognition. Second, the softmax output is essentially a point estimate of the multinomial distribution over the categorical probabilities so that softmax cannot capture the uncertainty of categorical probabilities, i.e., second-order uncertainty.

To overcome these limitations, recent evidential deep learning (EDL)~\cite{SensoyNIPS2018} is developed from the evidence framework of Dempster-Shafer Theory (DST)~\cite{SentzBook2002} and the subjective logic (SL)~\cite{JosangBook2016}. For a $K$-class classification problem, the EDL treats the input $\mathbf{x}$ as a proposition and regards the classification task as to give a multinomial subjective opinion in a $K$-dimensional domain $\{1,\ldots,K\}$. The subjective opinion is expressed as a triplet $\omega = (\mathbf{b}, u, \mathbf{a})$, where $\mathbf{b}=\{b_1,\ldots,b_K\}$ is the belief mass, $u$ represents the uncertainty, and $\mathbf{a}=\{a_1,\ldots,a_K\}$ is the base rate distribution. For any $k\in[1,2,\ldots,K]$, the probability mass of a multinomial opinion is defined as
\begin{equation}
    p_k = b_k + a_k u, \quad \forall y \in \mathbb{Y}
\label{pk}
\end{equation}
To enable the probability meaning of $p_k$, i.e., $\sum_k p_k=1$, the base rate $a_k$ is typically set to $1/K$ and the subjective opinion is constrained by
\begin{equation}
    u + \sum_{k=1}^{K} b_k = 1
\label{eq:ub}
\end{equation}

Besides, for a $K$-class setting, the probability mass $\mathbf{p}\!=\![p_1,p_2,\ldots,p_K]$ is assumed to follow a Dirichlet distribution parameterised by a $K$-dimensional Dirichlet strength vector $\boldsymbol{\alpha}=\{\alpha_1,\ldots,\alpha_K\}$:
\begin{equation}
\text{Dir}(\mathbf{p}|\boldsymbol{\alpha}) = \left\{ 
\begin{aligned}
& \frac{1}{B(\boldsymbol{\alpha})} \prod_{k=1}^{K} p_k^{\alpha_k-1}, & \text{for} \; \mathbf{p} \in \mathcal{S}_K, \\
& 0, & \text{otherwise,}
\end{aligned}
\right.
\end{equation}
where $B(\boldsymbol{\alpha})$ is a $K$-dimensional Beta function, $\mathcal{S}_K$ is a $K$-dimensional unit simplex. The total strength of the Dirichlet is defined as $S\!=\!\sum_{k=1}^{K}\alpha_k$. Note that for the special case when $K=2$, the Dirichlet distribution reduces to a Beta distribution and a binomial subjective opinion will be formulated in this case.

According to the evidence theory, the term \textit{evidence} is introduced to describe the amount of supporting observations for classifying the data $\mathbf{x}$ into a class. Let $\mathbf{e}=\{e_1,\ldots,e_K\}$ be the evidence for $K$ classes. Each entry $e_k\geq 0$ and the Dirichlet strength $\boldsymbol{\alpha}$ are linked according to the evidence theory by the following identity:
\begin{equation}
    \boldsymbol{\alpha} = \mathbf{e} + \mathbf{a} W
\label{eq:alpha}
\end{equation}
where $W$ is the weight of uncertain evidence. With the Dirichlet assumption, the expectation of the multinomial probability $\mathbf{p}$ is given by
\begin{equation}
    \mathbb{E}(p_k) = \frac{\alpha_k}{\sum_{k=1}^{K} \alpha_k} = \frac{e_k + a_k W}{W + \sum_{k=1}^K e_k}
\label{eq:expect}
\end{equation}
With loss of generality, the weight $W$ is set to $K$ and considering the assumption of the subjective opinion constraint in Eq.~\eqref{eq:ub} that $a_k=1/K$, we have the Dirichlet strength $\alpha_k=e_k+1$ according to Eq.~\eqref{eq:alpha}. In this way, the Dirichlet evidence can be mapped to the subjective opinion by setting the following equality's:
\begin{equation}
    b_k = \frac{e_k}{S} \quad \text{and} \quad u = \frac{K}{S} 
\label{bu}
\end{equation}
Therefore, we can see that if the evidence $e_k$ for the $k$-th class is predicted, the corresponding expected class probability in Eq. \eqref{pk} (or Eq.~\eqref{eq:expect}) can be rewritten as $p_k=\alpha_k / S$. From Eq. ~\eqref{bu}, it is clear that the predictive uncertainty $u$ can be determined after $\alpha_k$ is obtained. 

Inspired by this idea, the EDL leverages deep neural networks (DNN) to directly predict the evidence $\mathbf{e}$ from the given data $\mathbf{x}$ for a $K$-class classification problem. In particular, the output of the DNN is activated by a non-negative evidence function. Considering the Dirichlet prior, the DNN is trained by minimizing the negative log-likelihood:
\begin{equation}
\begin{split}
    \mathcal{L}_{EDL}^{(i)}(\mathbf{y}, \mathbf{e}; \theta) &= -\log \left(\int \prod_{k=1}^K p_{ik}^{y_{ik}} \frac{1}{B(\boldsymbol{\alpha}_{i})} \prod_{k=1}^{K} p_{ik}^{\alpha_{ik}-1} d\mathbf{p}_i\right) \\ 
    & = \sum_{k=1}^K y_{ik} \left(\log (S_i) - \log (e_{ik} + 1) \right)
\end{split}
\label{eq:supp_edl}
\end{equation}
where $\mathbf{y}_i=\{y_{i1},\ldots,y_{iK}\}$ is an one-hot $K$-dimensional label for sample $i$ and $\mathbf{e}_i$ can be expressed as $\mathbf{e}_i\!=\!g\left(f(\mathbf{x}_i;\theta)\right)$. Here, $f$ is the DNN parameterized by $\theta$ and $g$ is the evidence function such as $\exp$, softplus, or ReLU. Note that in~\cite{SensoyNIPS2018}, there are two other forms of EDL loss function. In our main paper, we found the Eq.~\eqref{eq:supp_edl} achieves better training empirical performance.

\subsection{EDL for Open Set Action Recognition}

To implement the EDL method on video action recognition tasks, we removed the Kullback–Leibler (KL) divergence regularizer term defined in~\cite{SensoyNIPS2018}, because the digamma function involved in the KL divergence is not numerically stable for large-scale video data. Instead, to compensate for the over-fitting risk, we propose the Evidential Uncertainty Calibration (EUC) as a new regularization. Together with the Contrastive Evidence Debiasing module, the complete training objective of our DEAR method can be expressed as
\begin{equation}
    \mathcal{L} = \sum_i \mathcal{L}_{EDL}^{(i)} + w_1 \mathcal{L}_{EUC} + w_2 \mathcal{L}_{CED}
\end{equation}
where $\mathcal{L}_{EUC}$ is defined in Eq.~\eqref{eq:loss_euc} in our main paper, and $\mathcal{L}_{CED}$ is the sum of (or one of for alternative training) $\mathcal{L}(\theta_f, \phi_f)$ and $\mathcal{L}(\theta_h, \phi_h)$ defined in Eq.~\eqref{debias} and Eq.~\eqref{bias} respectively in our main paper. The hyperparameters $w_1$ and $w_2$ are set to 1.0 and 0.1, respectively.

During the training process, the DEAR model aims to accurately construct the Dirichlet parameters $\boldsymbol{\alpha}$ by collecting the \textit{evidence} from human action video training set. In the inference phase, the probability of each action class is predicted as $\hat{p}_k=\alpha_k/S$ while the predictive uncertainty is simultaneously computed as $u=K/S$. If an input action video is assigned with high uncertainty, which means a vacuity of evidence to support for closed-set classification, the action is likely to be unknown from the open testing set.

Compared with existing DNN-based uncertainty estimation method such as Bayesian neural networks (BNN) or deep Gaussian process (DGP), the advantage of EDL is that the predictive uncertainty is deterministically learned without inexact posterior approximation and computationally expensive sampling. These merits enable the EDL method to be efficient for training recognition models from large-scale vision data such as the human action videos.

\subsection{Hilbert-Schmidt Independence Criterion}
\label{sec:hsic}

Hilbert-Schmidt Independence Criterion (HSIC) is a commonly-used dependency measurement of two high-dimensional variables. In practice, we used the unbiased HSIC estimator in~\cite{SongJMLR2012} with $m$ samples:
\begin{equation}
\resizebox{0.98\hsize}{!}{
    $\text{HSIC}^{k,l}(U,V)=\frac{1}{m(m-3)}\left[\text{tr}(\Tilde{U}\Tilde{V}^T) + \frac{\mathbf{1}^T\Tilde{U}\mathbf{1}\mathbf{1}^T\Tilde{V}\mathbf{1}}{(m-1)(m-2)} - \frac{2}{m-2}\mathbf{1}^T\Tilde{U}\Tilde{V}^T\mathbf{1}\right]$,
}
\end{equation}
where $\Tilde{U}$ is the kernelized matrix of $U$ with RBF kernel $k$ by $\Tilde{U}_{ij}=(1-\delta_{ij})k(u_i,u_j)$, $\{u_i\}\sim\!U$ and the $(1-\delta_{ij})$ sets the diagonal of $\Tilde{U}$ to zeros. $\Tilde{V}$ is defined similarly with kernel $l$, and $\mathbf{1}$ is an all-one vector. The HSIC value is equal to zero if and only if the two variables are independent.

\subsection{Evaluation of Model Calibration}

In our main paper, we used the expected calibration error (ECE) to quantitatively evaluate the model calibration performance of our proposed EUC method. According to~\cite{NaeiniAAAI2015,GuoICML2017}, the basic idea of model calibration is that, if the confidence estimation $\hat{p}$ (probability of correctness) is well calibrated, we hope $\hat{p}$ represent the true probability of the case when the predicted label $\hat{y}$ is correct. Formally, this can be expressed as
\begin{equation}
    \mathbb{P}(\hat{y}=y|\hat{p}=p) = p
\end{equation}
Since perfect calibration is infeasible due to the finite sample space, a practical way is to group all predicted confidence $\hat{p}$ into $M$ bins in the range of [0,1] such that the width of each bin is $1/M$. Therefore, for the $m$-th bin, the accuracy can be estimated by
\begin{equation}
    \text{acc}(B_m) = \frac{1}{|B_m|}\sum_{i\in B_m}\mathbb{I}(\hat{y}_i = y_i)
\end{equation}
where $B_m$ is the set of indices of prediction $\hat{p}$ when it falls into the $m$-th bin. $\hat{y}_i$ and $y_i$ are predicted and ground truth labels. Besides, the average confidence for the $m$-th bin can be expressed as
\begin{equation}
    \text{conf}(B_m) = \frac{1}{|B_m|}\sum_{i\in B_m} \hat{p}_i
\end{equation}
To evaluate the mis-calibration error, the ECE is defined as the expectation of the gap between the accuracy and confidence in $M$ bins for all $N$ samples:
\begin{equation}
    \text{ECE} = \sum_{m=1}^{M} \frac{|B_m|}{N} |\text{acc}(B_m) - \text{conf}(B_m)|
\end{equation}
A perfect calibrated model means that ECE=0 and higher ECE value indicates that the model is less calibrated.

\section{Implementation Details}
\label{implem}

\textbf{Network Architecture.} As presented in our main paper, the proposed DEAR method as well as all other baselines are implemented on top of the four recent video action recognition models, i.e., I3D, TSM, SlowFast, and TPN. For simplicity, these models  use ResNet-50 as the backbone architecture and the network weights are initialized with the pre-trained model from the Kinetics-400 benchmark. To avoid the impact of the validation experiments on the Kinetics and Mimetics datasets, the pre-trained model is not used and we train the model from scratch using the same hyperparameters.

Specifically, for the \textbf{I3D} model, it is straightforward to implement our method by replacing the cross-entropy loss with the proposed EUC regularized EDL loss, and inserting the proposed CED module before the recognition head (fully-connected layers). For the \textbf{TSM} model, since the architecture of TSM is based on 2D convolution where the output feature embedding is with the size $(B, MC, H, W)$, we recover the number of video segments $M$ as the temporal dimension such that the 5-dimensional tensor with size $(B, C, M, H, W)$ could be compatible with our proposed CED module for contrastive debiasing. For the \textbf{SlowFast} model, our CED module is inserted after the \textit{slow} pathway because the feature embedding of slow pathway is more likely to be biased since it captures the static cues of video content. For the \textbf{TPN} model, we used the ResNet-50-like SlowOnly model as the recognition backbone and the auxiliary cross-entropy loss in the TPN head is kept unchanged.

\textbf{Training and Inference.} In the training phase, we choose the $\exp$ function as the evidence function because we empirically found $\exp$ is numerically more stable when using the proposed EDL loss $\mathcal{L}_{EDL}$. We set the hyperparameter $\lambda_0$ to 0.01 in EUC loss $\mathcal{L}_{EUC}$ and set $\lambda$ to 1.0 in the two CED losses. The weight of $\mathcal{L}_{EUC}$ is set to 1.0 and the weight of the sum of the two CED losses is empirically set to 0.1. In practice, we found the model performance is robust to these hyperparameters. We used mini-batch SGD with nesterov strategy to train all the 3D convolution models. For all models, weight decay is set to 0.0001 and momentum factor is set to 0.9 by default. Our experiments are supported by two GeoForce RTX 3090 and two Tesla A100 GPUs. Since no additional parameters are introduced during inference, the inference speed of existing action recognition models is not affected.

\textbf{Dataset Information.} For the UCF-101 and HMDB-51 datasets, we used the \textit{split1} for all experiments. For the MiT-v2 dataset, we only use the testing set for evaluation. To validate the proposed CED module, we refer to~\cite{BahngICML2020} and select 10 action categories which are included in both Kinetics and Mimetics dataset. These categories are \textit{canoeing or kayaking}, \textit{climbing a rope}, \textit{driving car}, \textit{golf driving}, \textit{opening bottle}, \textit{playing piano}, \textit{playing volleyball}, \textit{shooting goal (soccer)}, \textit{surfing water}, and \textit{writing}. The recognition model is trained from scratch on the 10 categories of Kinetics training set, and tested on these categories of both Kinetics and Mimetics testing set.

\section{Quantitative Results}
\label{quanti}


\newcommand{\curvefigwidth}{0.32\textwidth}
\begin{figure*}[t]
    \centering
    \subcaptionbox{TSM}{
        \includegraphics[width=\curvefigwidth]{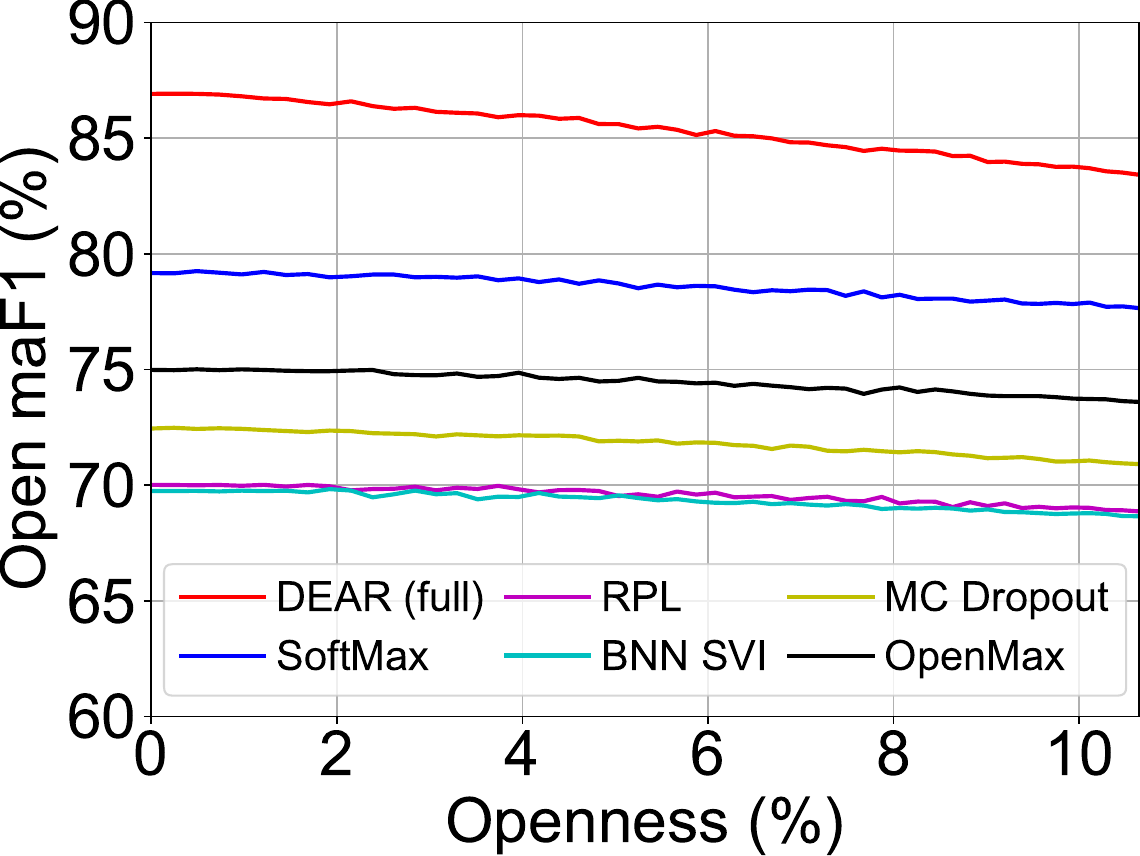}
    }
    \subcaptionbox{SlowFast}{
        \includegraphics[width=\curvefigwidth]{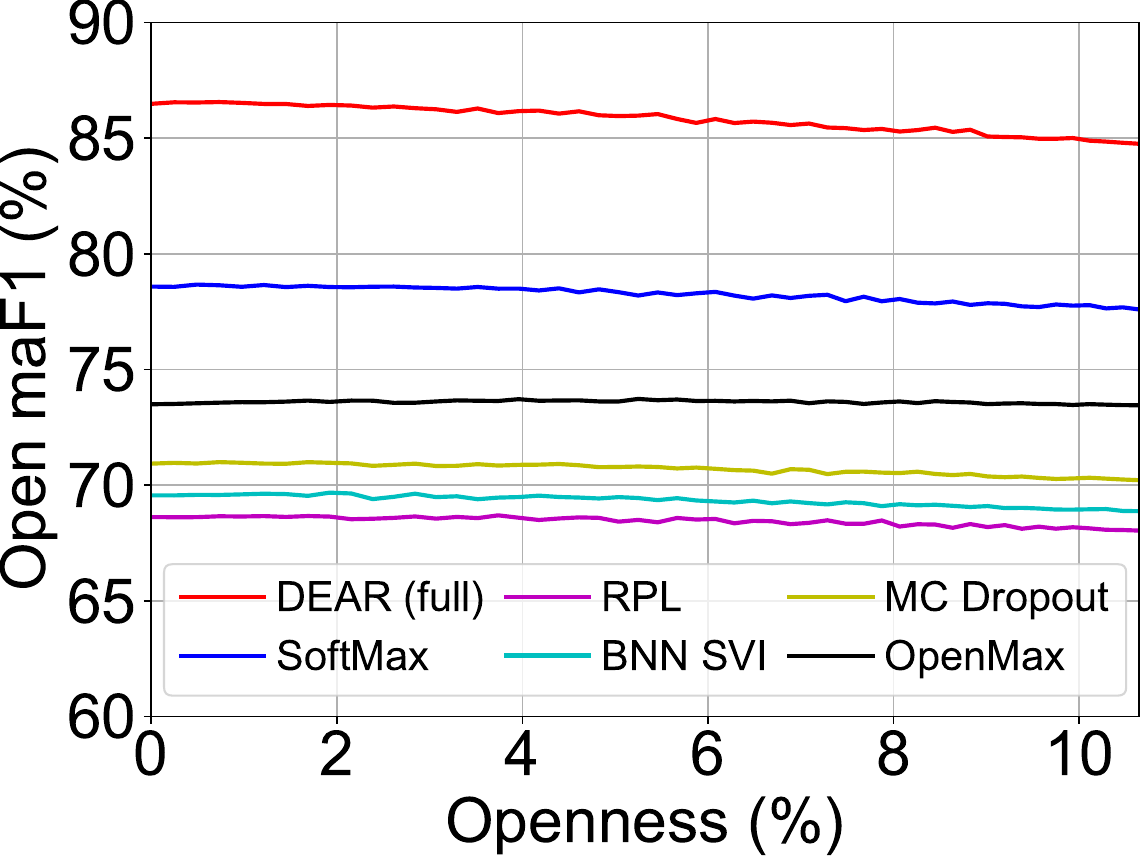}
    }
    \subcaptionbox{TPN}{
        \includegraphics[width=\curvefigwidth]{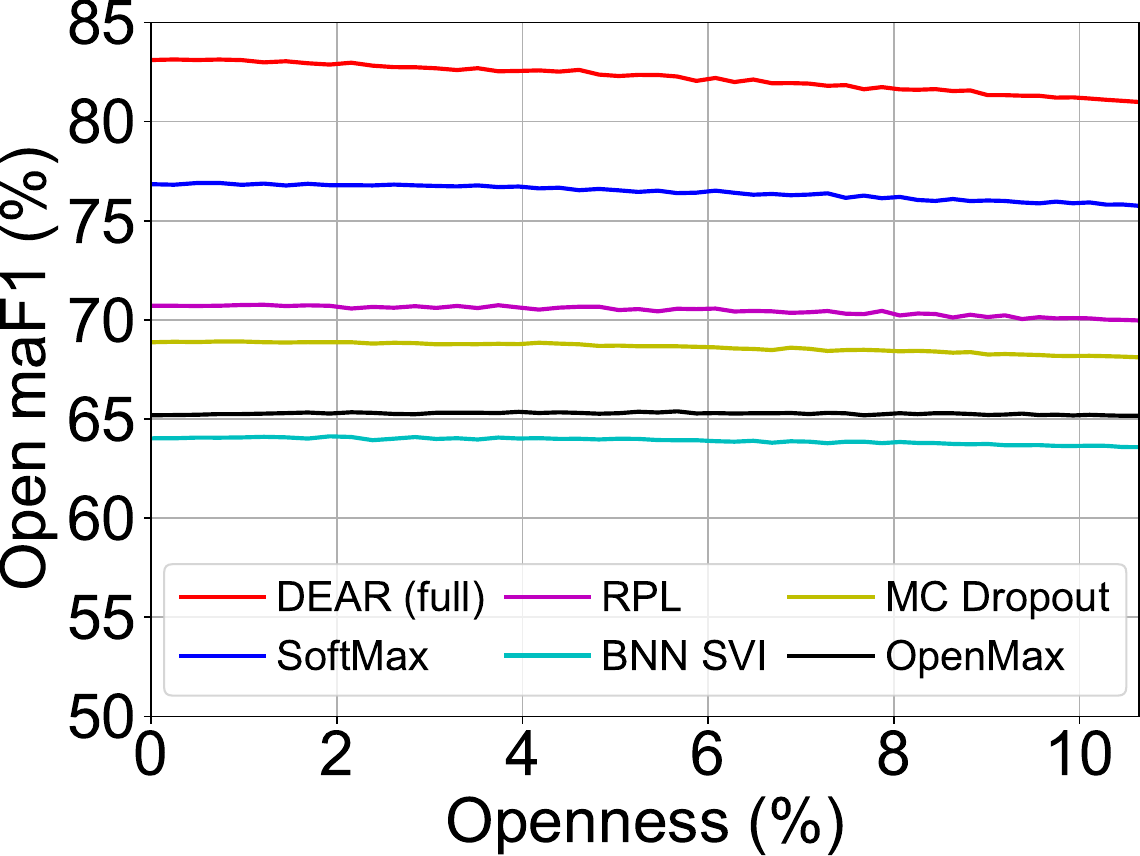}
    }
\caption{\textbf{Open macro-F1 scores against varying Openness}. The HMDB-51 testing set is used as the unknown.}
\label{fig:curve_hmdb}
\end{figure*}

\begin{figure*}[t]
    \centering
    \subcaptionbox{TSM}{
        \includegraphics[width=\curvefigwidth]{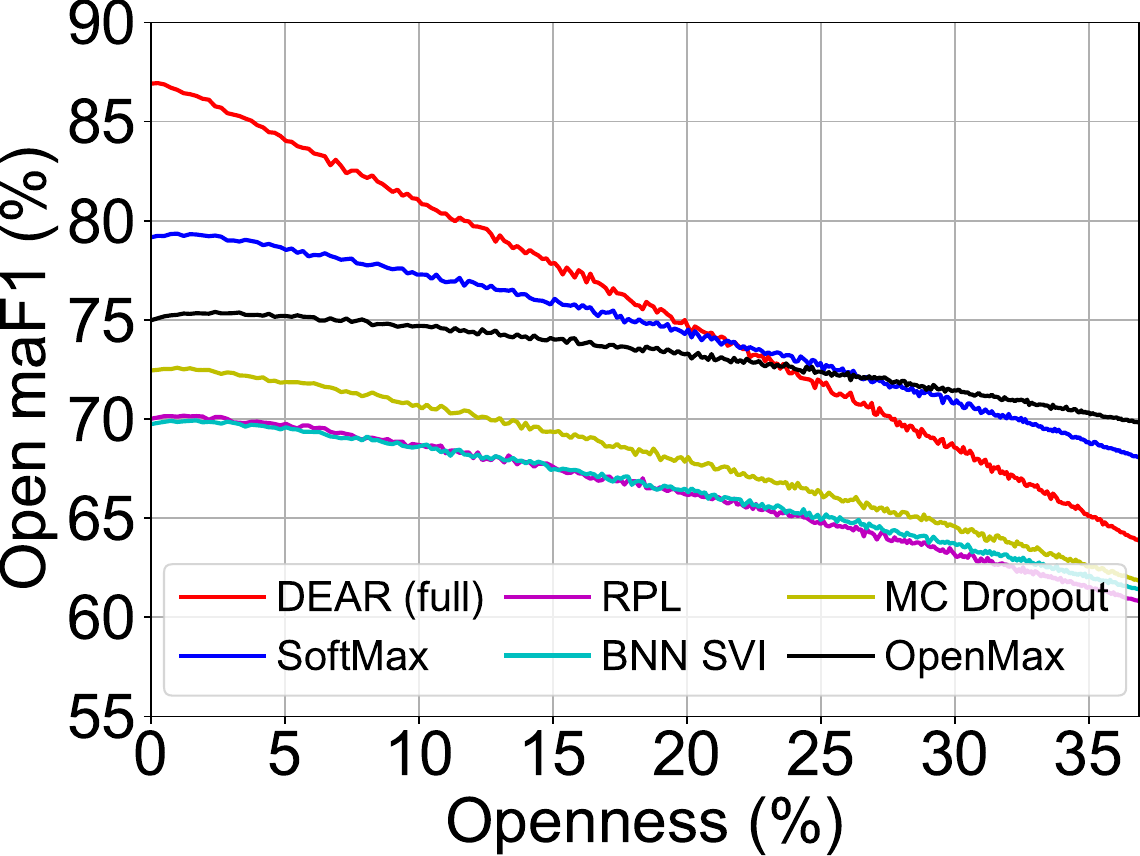}
    }
    \subcaptionbox{SlowFast}{
        \includegraphics[width=\curvefigwidth]{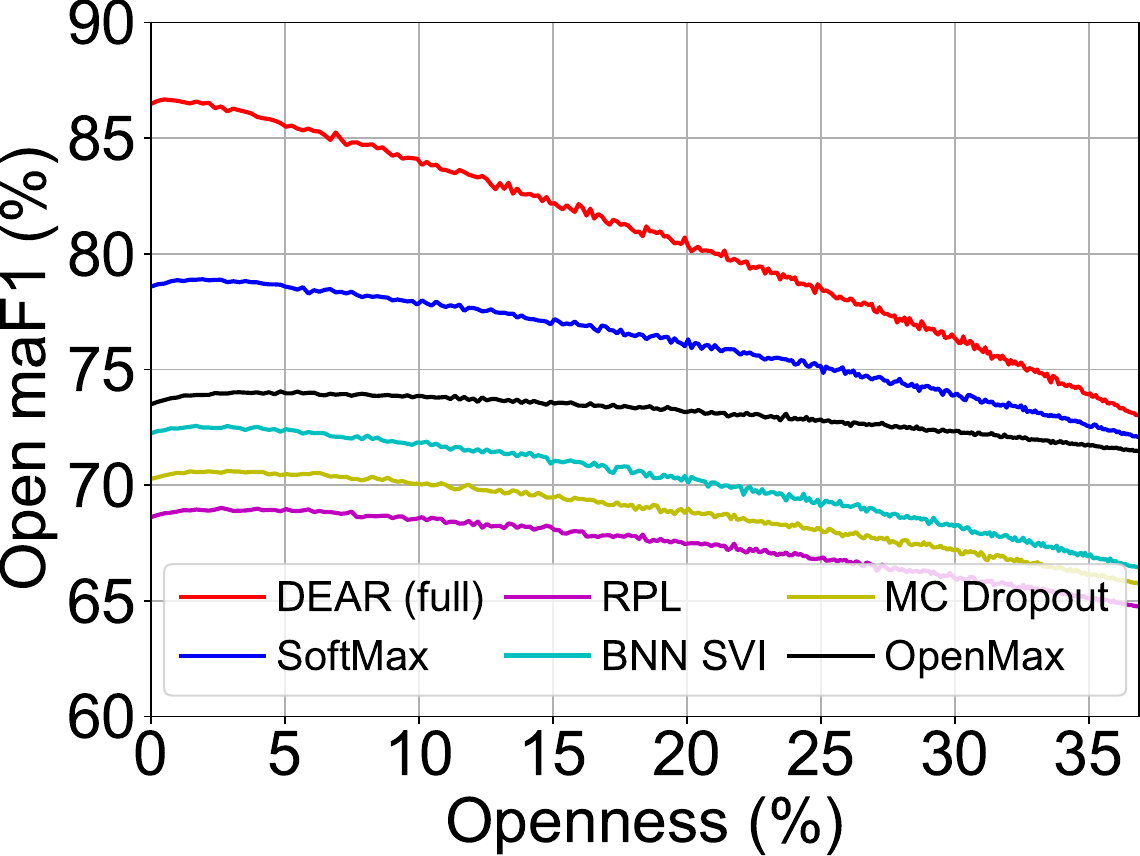}
    }
    \subcaptionbox{TPN}{
        \includegraphics[width=\curvefigwidth]{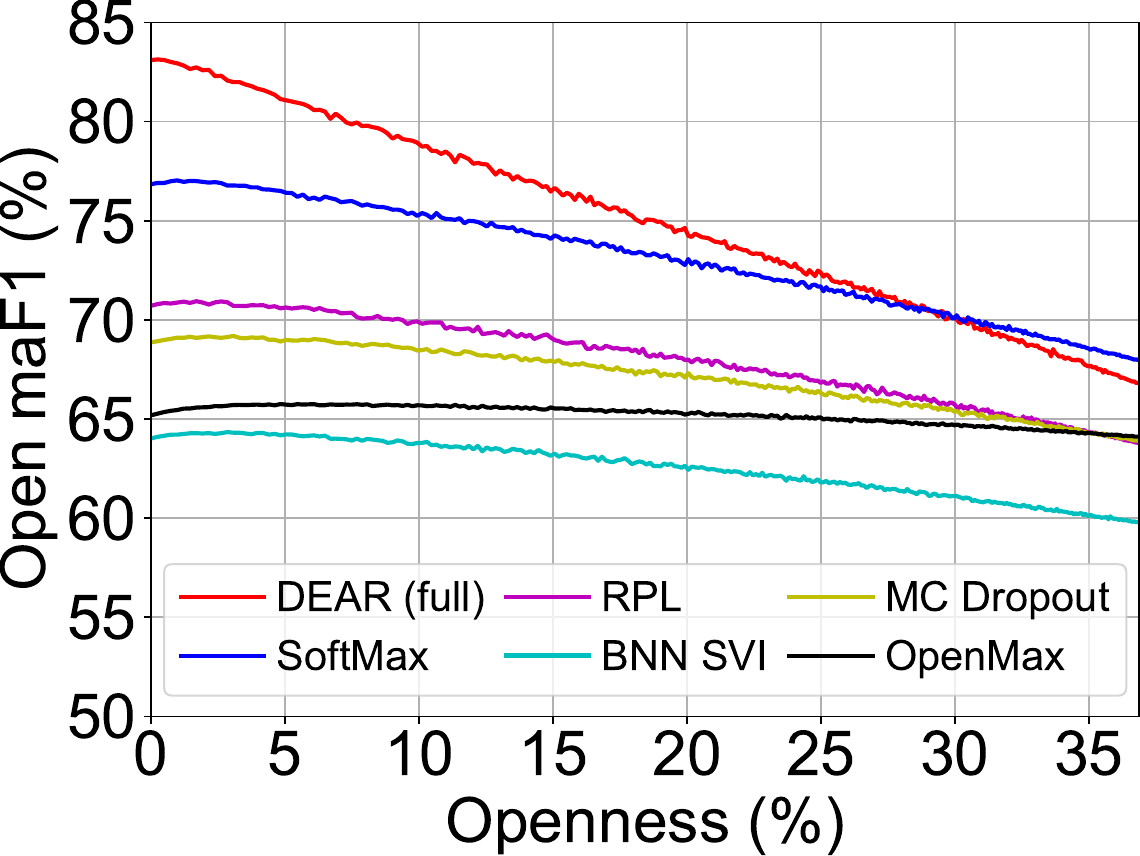}
    }
\caption{\textbf{Open macro-F1 scores against varying Openness}. The MiT-v2 testing set is used as unknown.}
\label{fig:curve_mit}
\end{figure*}

\textbf{Open Set Action Recognition}. In addition to the I3D-based curves of Open maF1 scores against varying openness in our main paper, we also provide the curves for other action recognition models, including TSM, SlowFast, and TPN in Fig.~\ref{fig:curve_hmdb} and Fig.~\ref{fig:curve_mit}. The figures show that when HMDB-51 testing set is used as the unknown, the proposed DEAR method significantly outperforms other baselines with large margins. When MiT-v2 testing set is used as the unknown, the DEAR method could achieve the best performance with relatively low openness.

\begin{figure*}[t]
    \centering
    \subcaptionbox{SoftMax}{
        \includegraphics[width=0.32\textwidth]{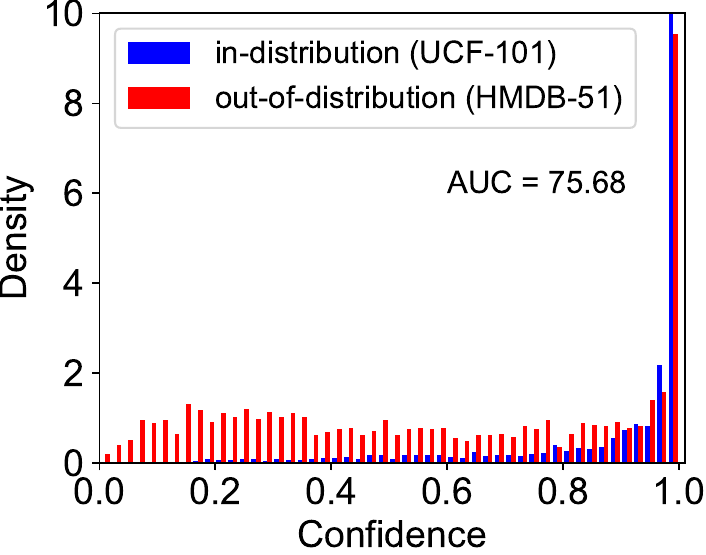}
    }
    \subcaptionbox{OpenMax~\cite{BendaleCVPR2016}}{
        \includegraphics[width=0.32\textwidth]{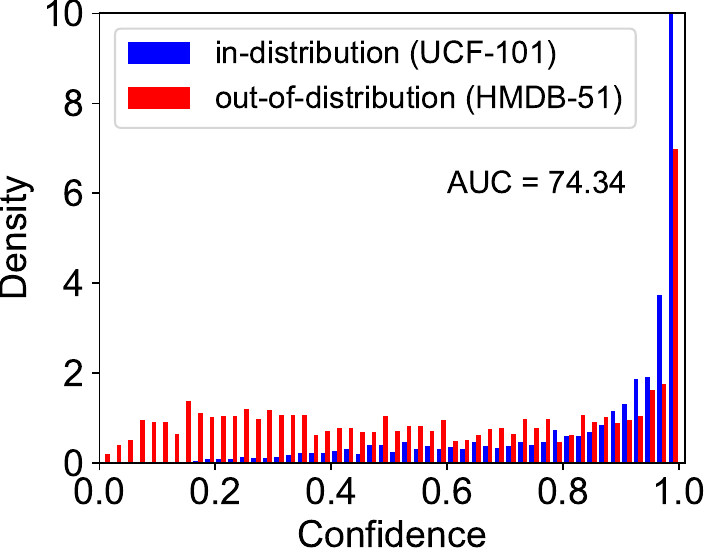}
    }
    \subcaptionbox{RPL~\cite{ChenECCV2020}}{
        \includegraphics[width=0.32\textwidth]{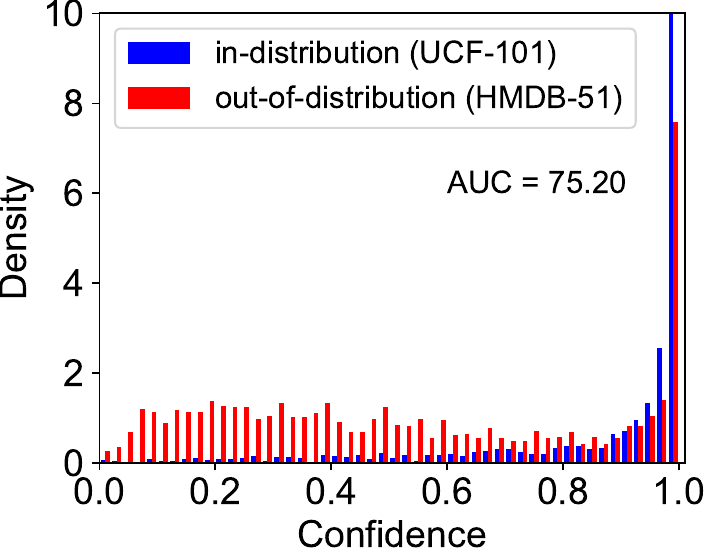}
    }
    \vfill
    \subcaptionbox{MC Dropout}{
        \includegraphics[width=0.32\textwidth]{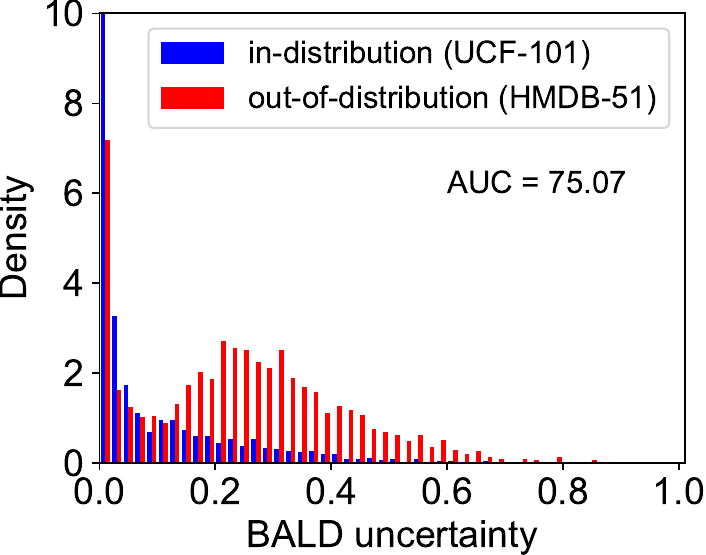}
    }
    \subcaptionbox{BNN SVI~\cite{KrishnanNIPS2018}}{
        \includegraphics[width=0.32\textwidth]{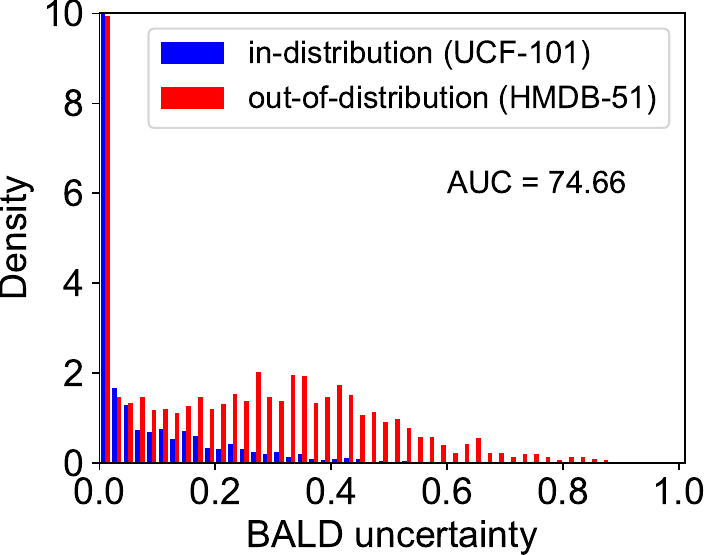}
    }
    \subcaptionbox{DEAR (full)}{
        \includegraphics[width=0.32\textwidth]{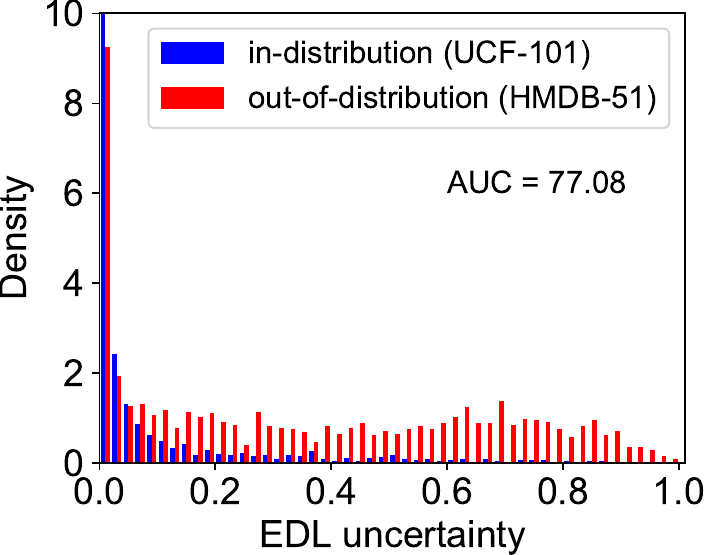}
    }
    \captionsetup{font=small,aboveskip=5pt}
    \caption{\textbf{I3D-based Out-of-distribution Detection with HMDB-51 as Unknown.} Values are normalized to [0,1] within each distribution.}
    \label{fig:ood_i3d_hmdb}
\end{figure*}
\begin{figure*}[t]
    \centering
    \subcaptionbox{SoftMax}{
        \includegraphics[width=0.32\textwidth]{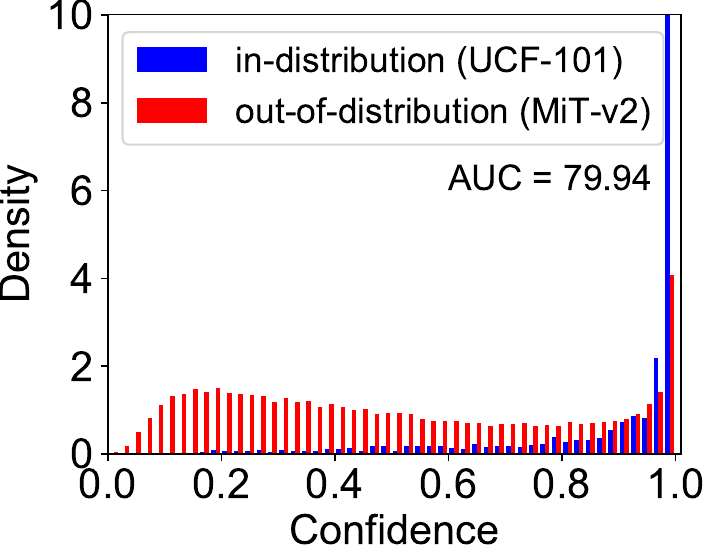}
    }
    \subcaptionbox{OpenMax~\cite{BendaleCVPR2016}}{
        \includegraphics[width=0.32\textwidth]{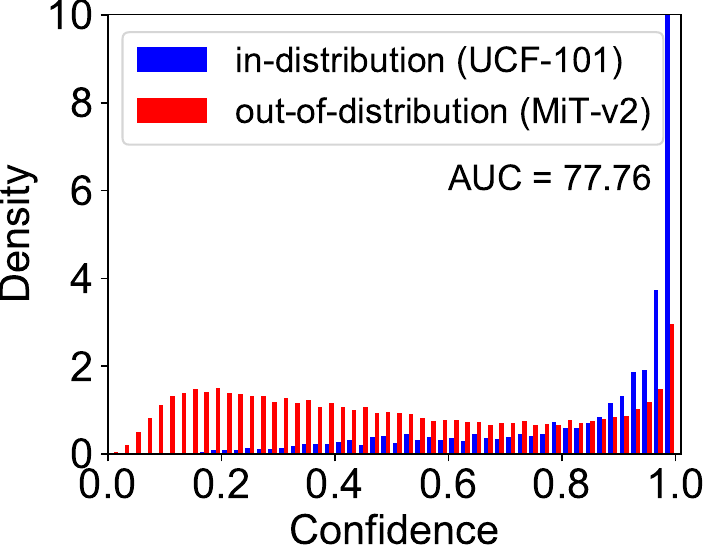}
    }
    \subcaptionbox{RPL~\cite{ChenECCV2020}}{
        \includegraphics[width=0.32\textwidth]{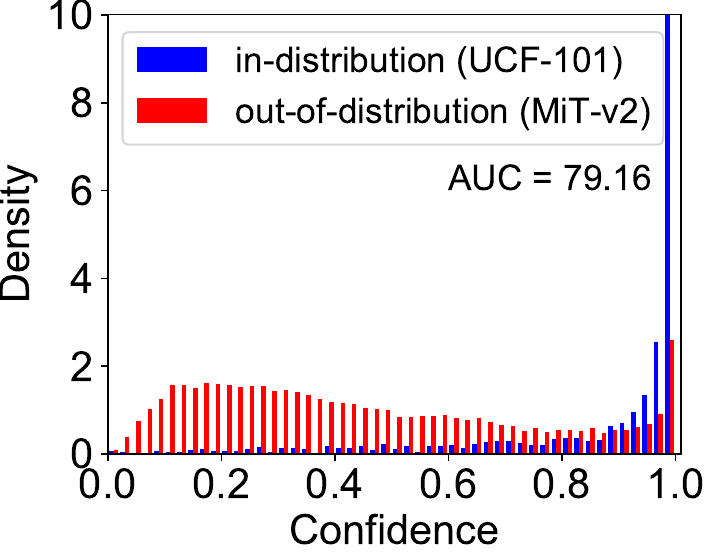}
    }
    \vfill
    \subcaptionbox{MC Dropout}{
        \includegraphics[width=0.32\textwidth]{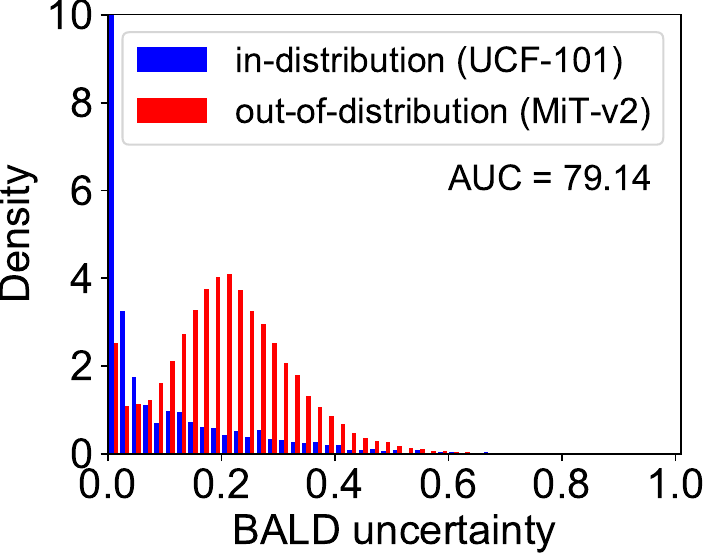}
    }
    \subcaptionbox{BNN SVI~\cite{KrishnanNIPS2018}}{
        \includegraphics[width=0.32\textwidth]{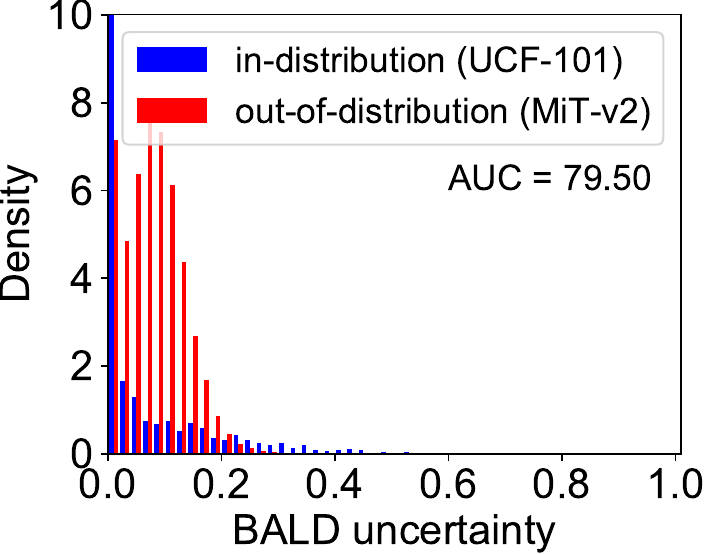}
    }
    \subcaptionbox{DEAR (full)}{
        \includegraphics[width=0.32\textwidth]{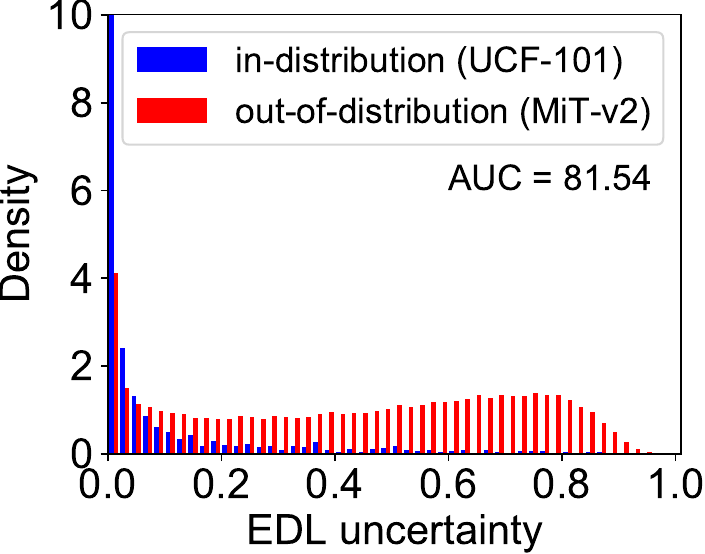}
    }
    \captionsetup{font=small,aboveskip=5pt}
    \caption{\textbf{I3D-based Out-of-distribution Detection with MiT-v2 as Unknown.} Values are normalized to [0,1] within each distribution.}
    \label{fig:ood_i3d_mit}
\end{figure*}
\begin{figure*}[t]
    \centering
    \subcaptionbox{SoftMax}{
        \includegraphics[width=0.32\textwidth]{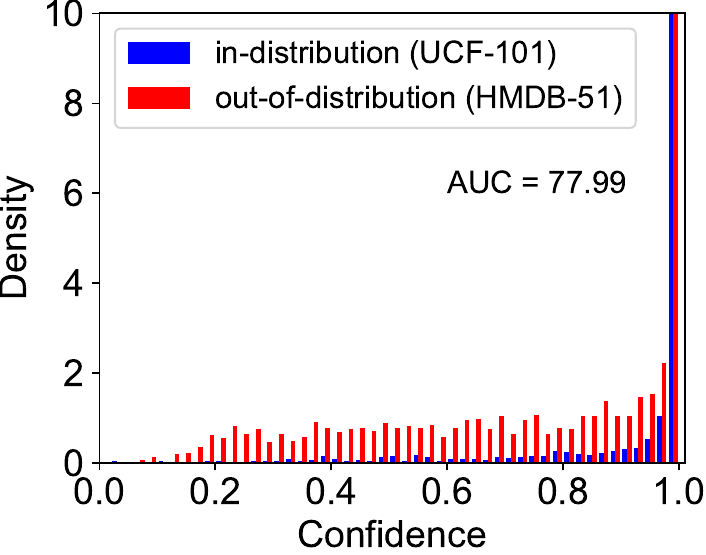}
    }
    \subcaptionbox{OpenMax~\cite{BendaleCVPR2016}}{
        \includegraphics[width=0.32\textwidth]{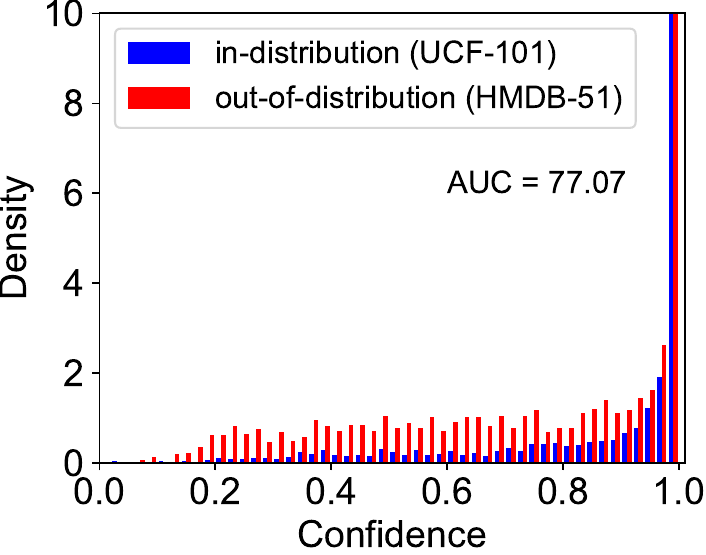}
    }
    \subcaptionbox{RPL~\cite{ChenECCV2020}}{
        \includegraphics[width=0.32\textwidth]{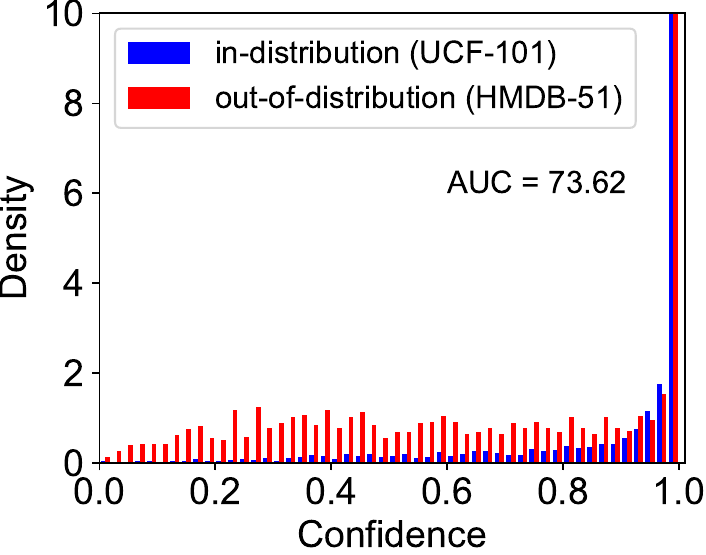}
    }
    \vfill
    \subcaptionbox{MC Dropout}{
        \includegraphics[width=0.32\textwidth]{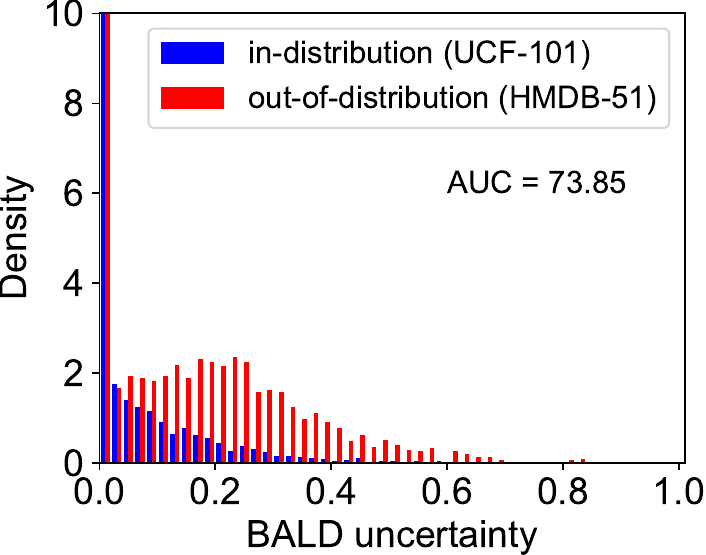}
    }
    \subcaptionbox{BNN SVI~\cite{KrishnanNIPS2018}}{
        \includegraphics[width=0.32\textwidth]{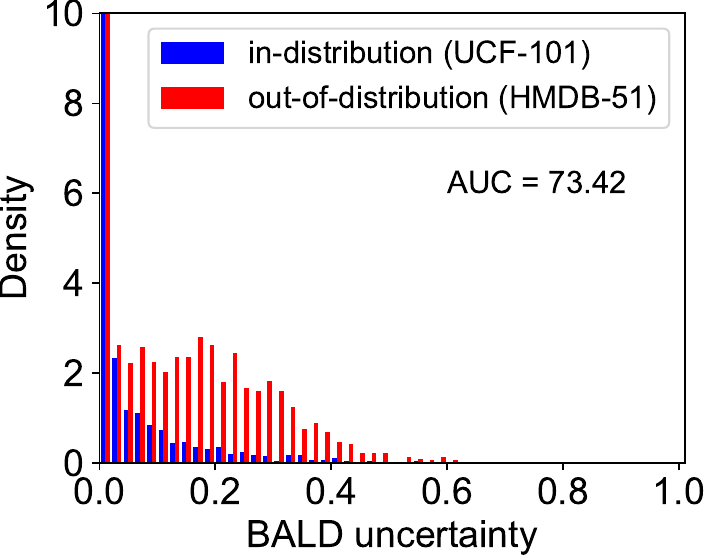}
    }
    \subcaptionbox{DEAR (full)}{
        \includegraphics[width=0.32\textwidth]{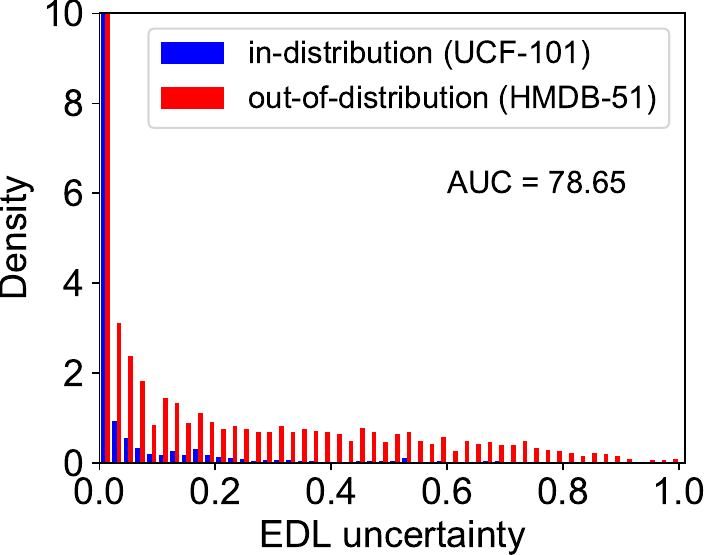}
    }
    \captionsetup{font=small,aboveskip=5pt}
    \caption{\textbf{TSM-based Out-of-distribution Detection with HMDB-51 as Unknown.} Values are normalized to [0,1] within each distribution.}
    \label{fig:ood_tsm_hmdb}
\end{figure*}
\begin{figure*}[t]
    \centering
    \subcaptionbox{SoftMax}{
        \includegraphics[width=0.32\textwidth]{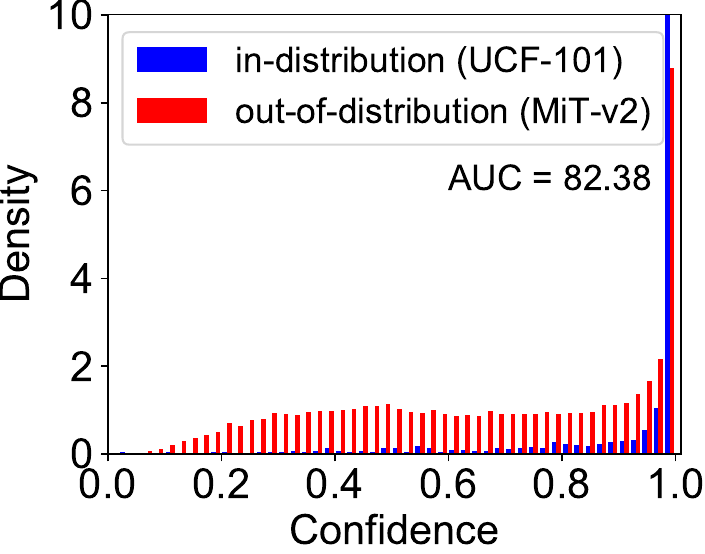}
    }
    \subcaptionbox{OpenMax~\cite{BendaleCVPR2016}}{
        \includegraphics[width=0.32\textwidth]{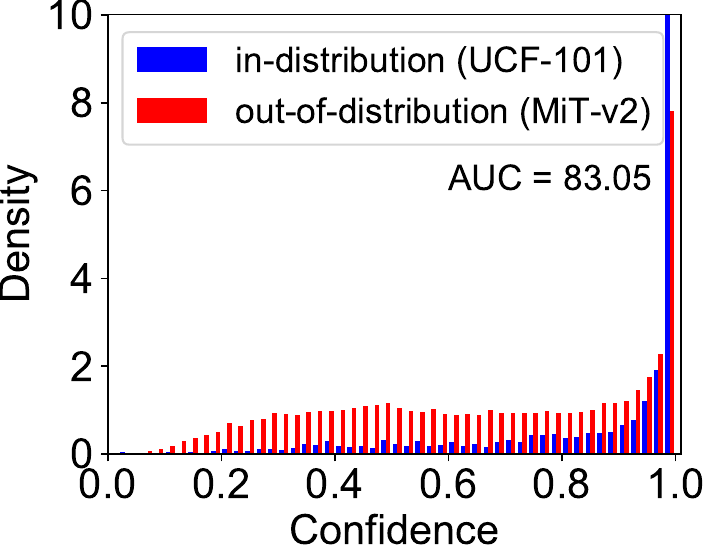}
    }
    \subcaptionbox{RPL~\cite{ChenECCV2020}}{
        \includegraphics[width=0.32\textwidth]{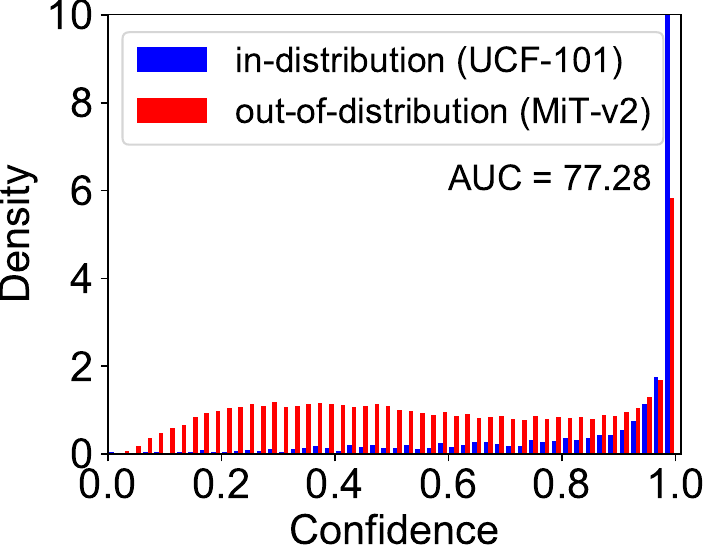}
    }
    \vfill
    \subcaptionbox{MC Dropout}{
        \includegraphics[width=0.32\textwidth]{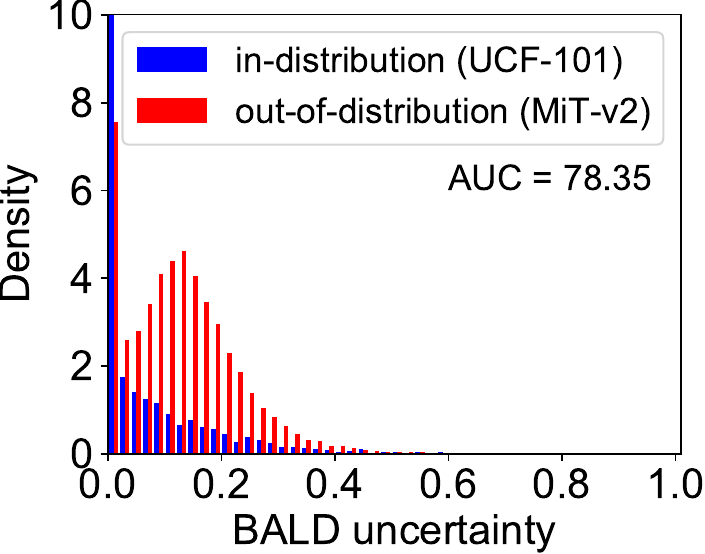}
    }
    \subcaptionbox{BNN SVI~\cite{KrishnanNIPS2018}}{
        \includegraphics[width=0.32\textwidth]{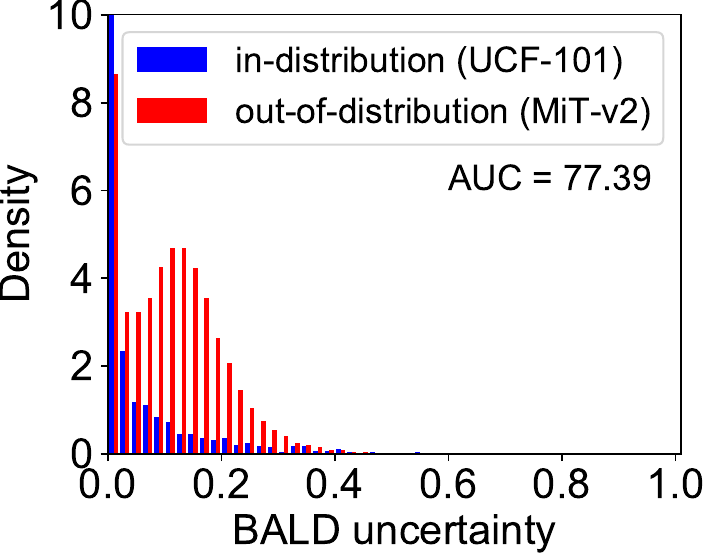}
    }
    \subcaptionbox{DEAR (full)}{
        \includegraphics[width=0.32\textwidth]{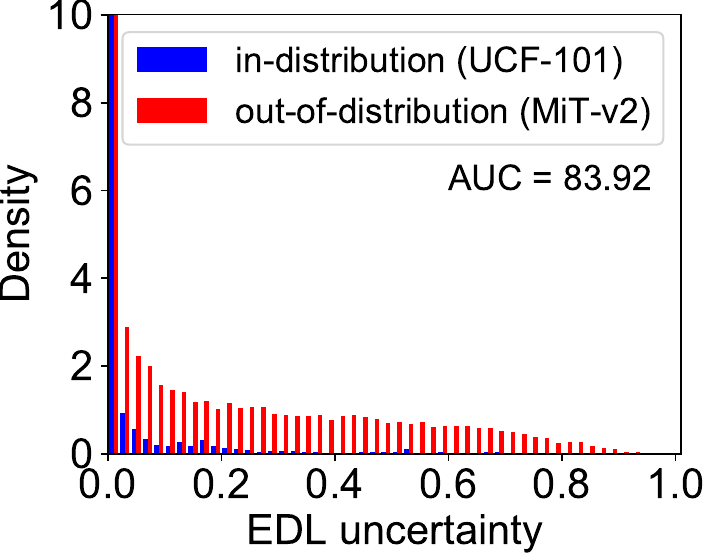}
    }
    \captionsetup{font=small,aboveskip=5pt}
    \caption{\textbf{TSM-based Out-of-distribution Detection with MiT-v2 as Unknown.} Values are normalized to [0,1] within each distribution.}
    \label{fig:ood_tsm_mit}
\end{figure*}
\begin{figure*}[t]
    \centering
    \subcaptionbox{SoftMax}{
        \includegraphics[width=0.32\textwidth]{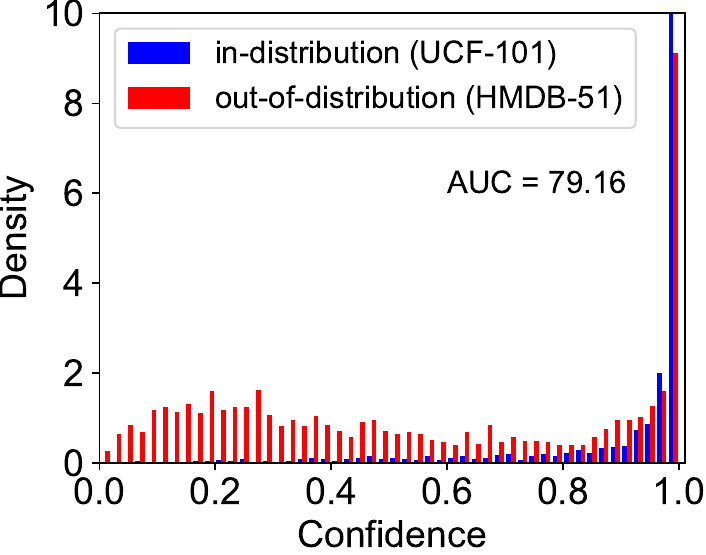}
    }
    \subcaptionbox{OpenMax~\cite{BendaleCVPR2016}}{
        \includegraphics[width=0.32\textwidth]{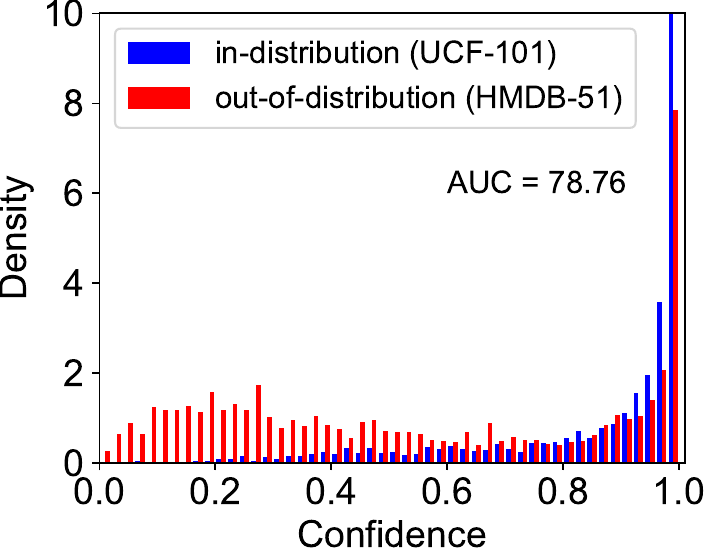}
    }
    \subcaptionbox{RPL~\cite{ChenECCV2020}}{
        \includegraphics[width=0.32\textwidth]{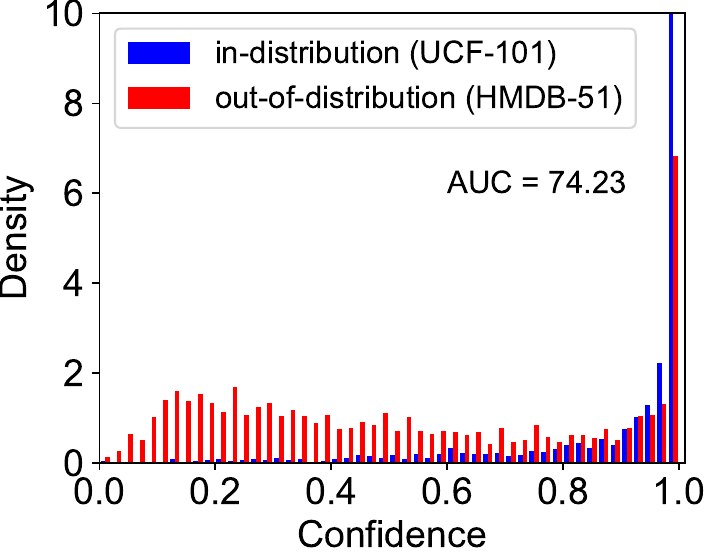}
    }
    \vfill
    \subcaptionbox{MC Dropout}{
        \includegraphics[width=0.32\textwidth]{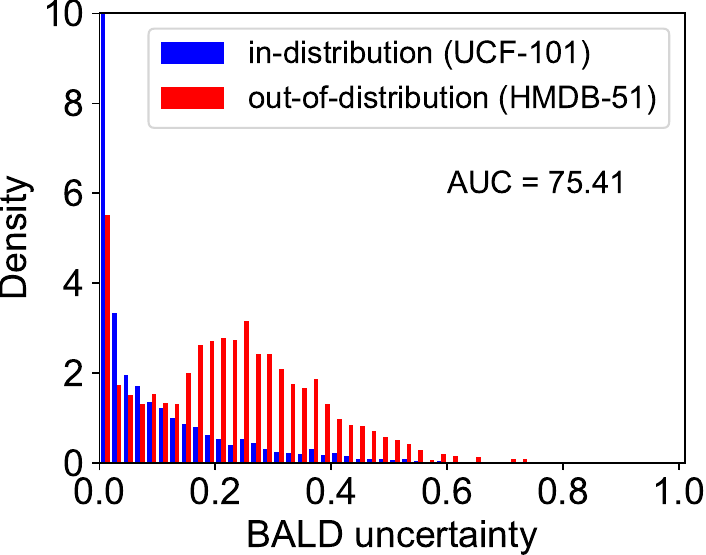}
    }
    \subcaptionbox{BNN SVI~\cite{KrishnanNIPS2018}}{
        \includegraphics[width=0.32\textwidth]{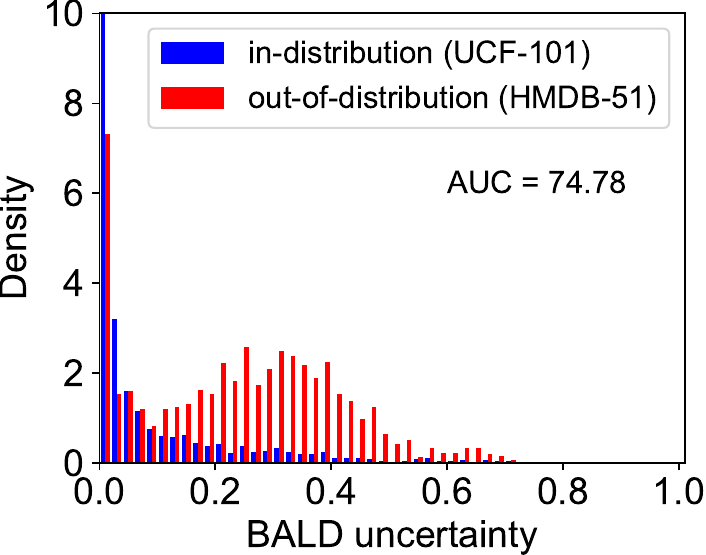}
    }
    \subcaptionbox{DEAR (full)}{
        \includegraphics[width=0.32\textwidth]{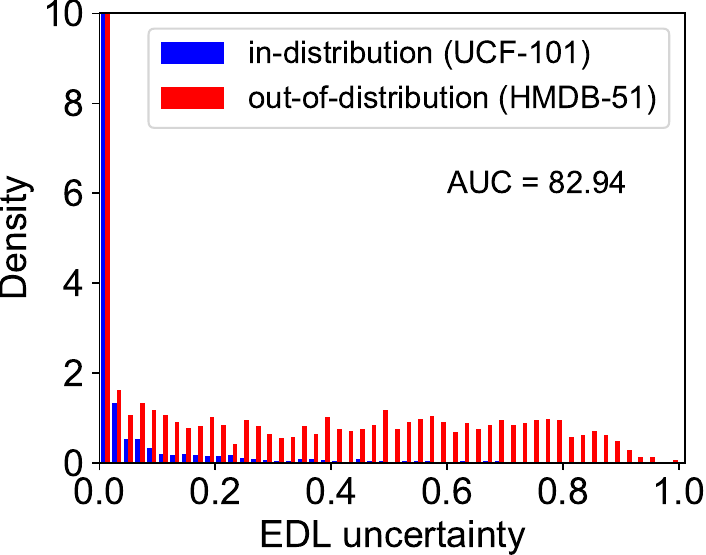}
    }
    \captionsetup{font=small,aboveskip=5pt}
    \caption{\textbf{SlowFast-based Out-of-distribution Detection with HMDB-51 as Unknown.} Values are normalized to [0,1] within each distribution.}
    \label{fig:ood_slowfast_hmdb}
\end{figure*}
\begin{figure*}[t]
    \centering
    \subcaptionbox{SoftMax}{
        \includegraphics[width=0.32\textwidth]{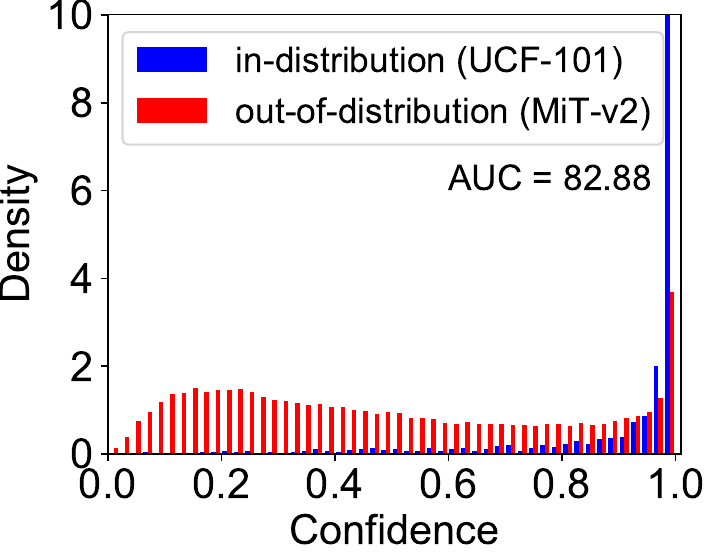}
    }
    \subcaptionbox{OpenMax~\cite{BendaleCVPR2016}}{
        \includegraphics[width=0.32\textwidth]{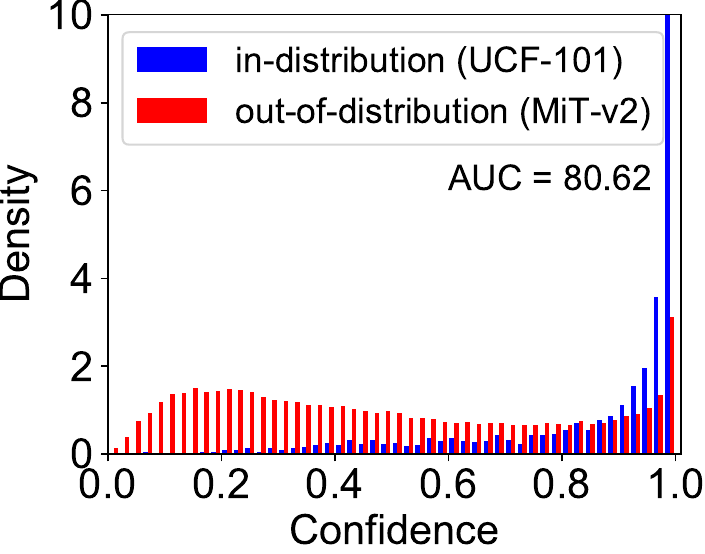}
    }
    \subcaptionbox{RPL~\cite{ChenECCV2020}}{
        \includegraphics[width=0.32\textwidth]{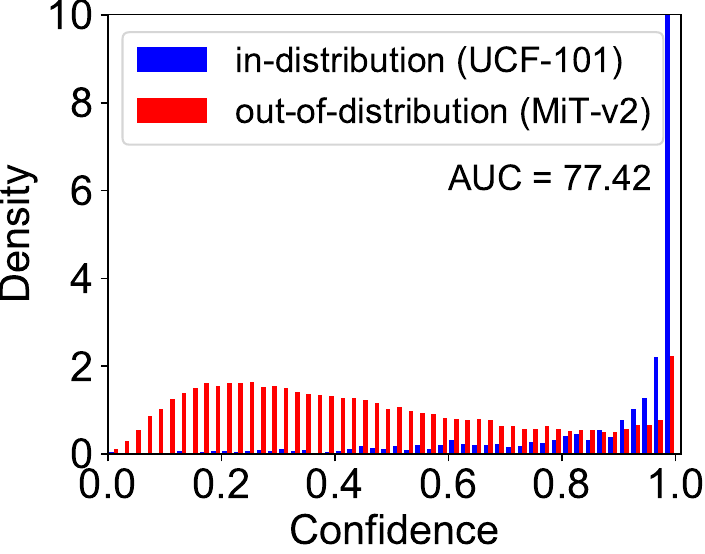}
    }
    \vfill
    \subcaptionbox{MC Dropout}{
        \includegraphics[width=0.32\textwidth]{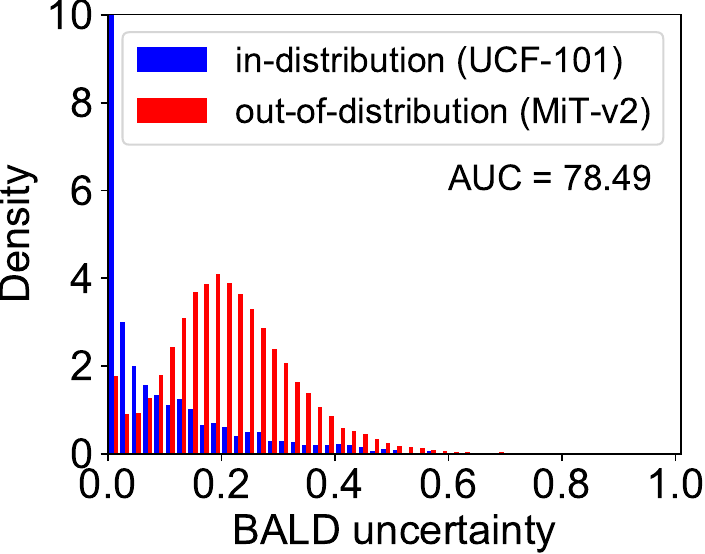}
    }
    \subcaptionbox{BNN SVI~\cite{KrishnanNIPS2018}}{
        \includegraphics[width=0.32\textwidth]{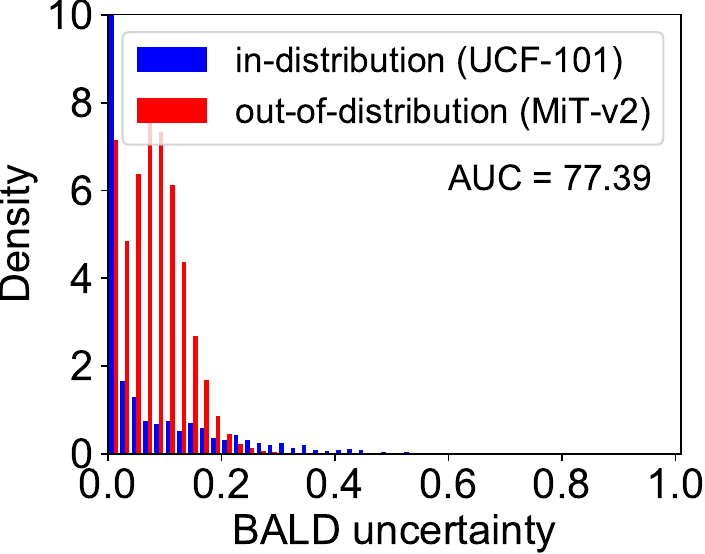}
    }
    \subcaptionbox{DEAR (full)}{
        \includegraphics[width=0.32\textwidth]{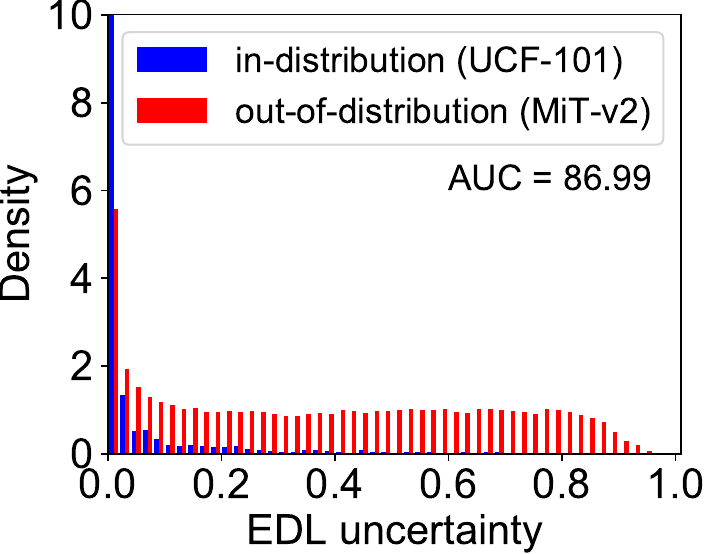}
    }
    \captionsetup{font=small,aboveskip=5pt}
    \caption{\textbf{SlowFast-based Out-of-distribution Detection with MiT-v2 as Unknown.} Values are normalized to [0,1] within each distribution.}
    \label{fig:ood_slowfast_mit}
\end{figure*}
\begin{figure*}[t]
    \centering
    \subcaptionbox{SoftMax}{
        \includegraphics[width=0.32\textwidth]{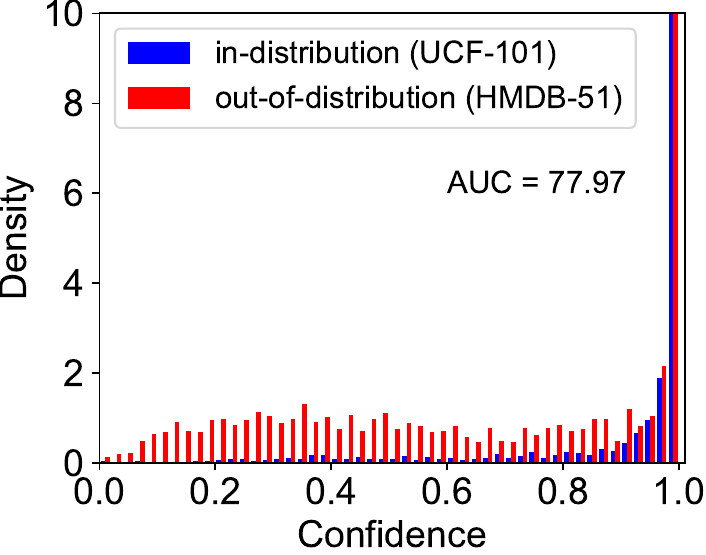}
    }
    \subcaptionbox{OpenMax~\cite{BendaleCVPR2016}}{
        \includegraphics[width=0.32\textwidth]{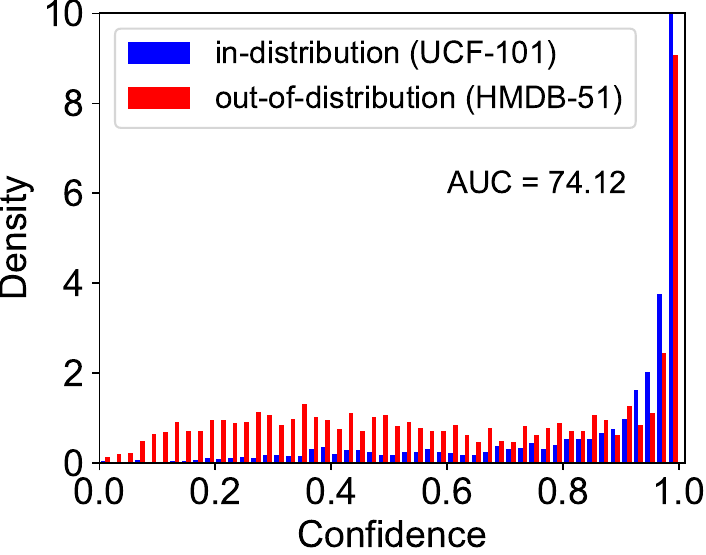}
    }
    \subcaptionbox{RPL~\cite{ChenECCV2020}}{
        \includegraphics[width=0.32\textwidth]{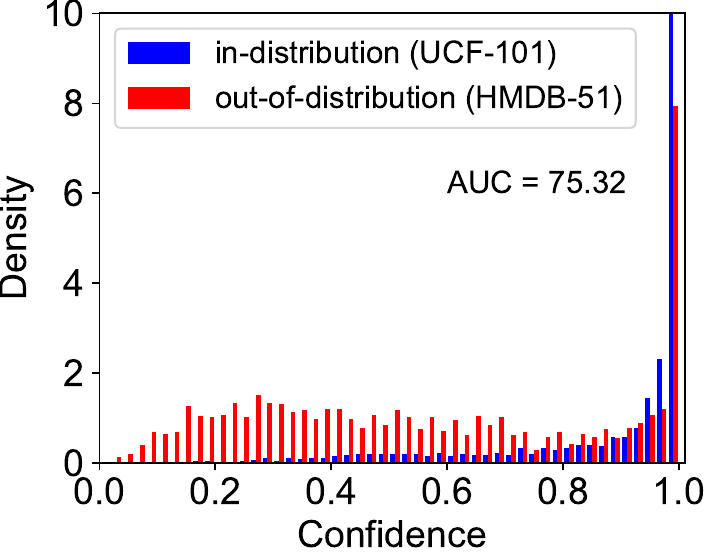}
    }
    \vfill
    \subcaptionbox{MC Dropout}{
        \includegraphics[width=0.32\textwidth]{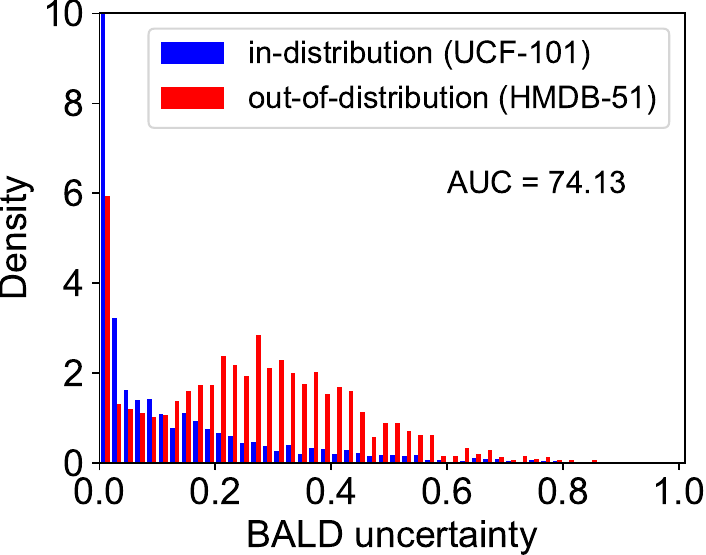}
    }
    \subcaptionbox{BNN SVI~\cite{KrishnanNIPS2018}}{
        \includegraphics[width=0.32\textwidth]{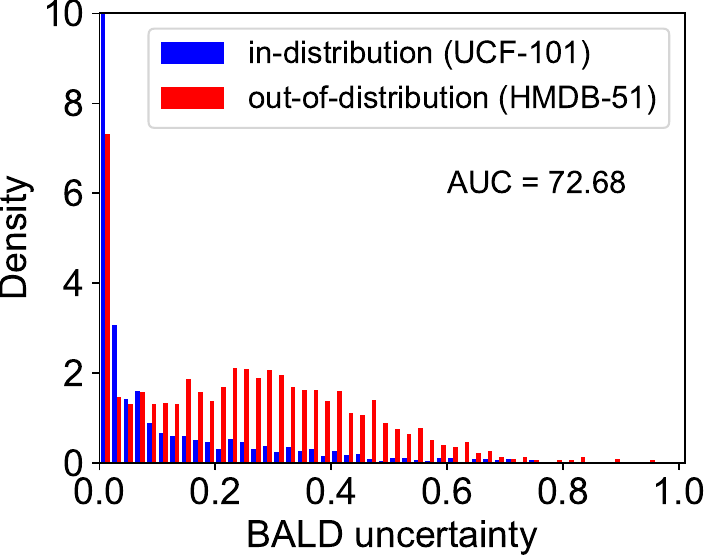}
    }
    \subcaptionbox{DEAR (full)}{
        \includegraphics[width=0.32\textwidth]{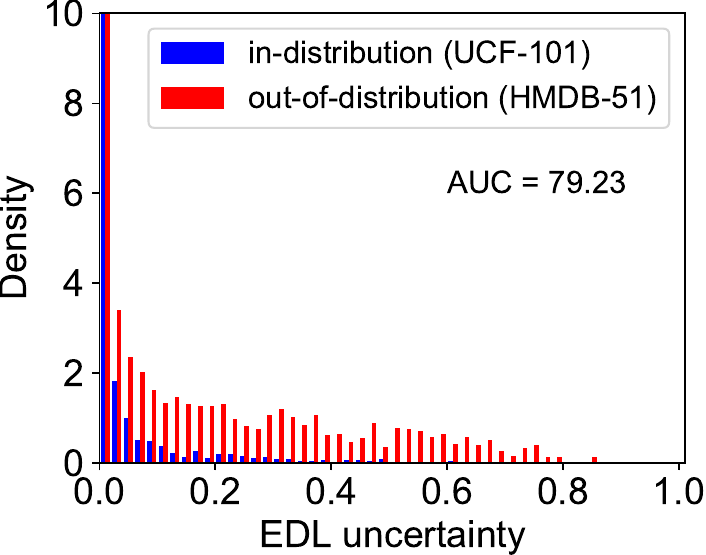}
    }
    \captionsetup{font=small,aboveskip=5pt}
    \caption{\textbf{TPN-based Out-of-distribution Detection with HMDB-51 as Unknown.} Values are normalized to [0,1] within each distribution.}
    \label{fig:ood_tpn_hmdb}
\end{figure*}
\begin{figure*}[t]
    \centering
    \subcaptionbox{SoftMax}{
        \includegraphics[width=0.32\textwidth]{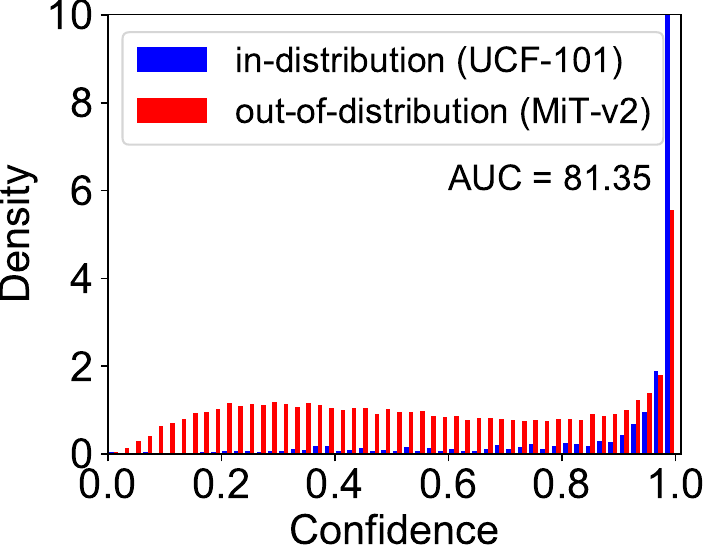}
    }
    \subcaptionbox{OpenMax~\cite{BendaleCVPR2016}}{
        \includegraphics[width=0.32\textwidth]{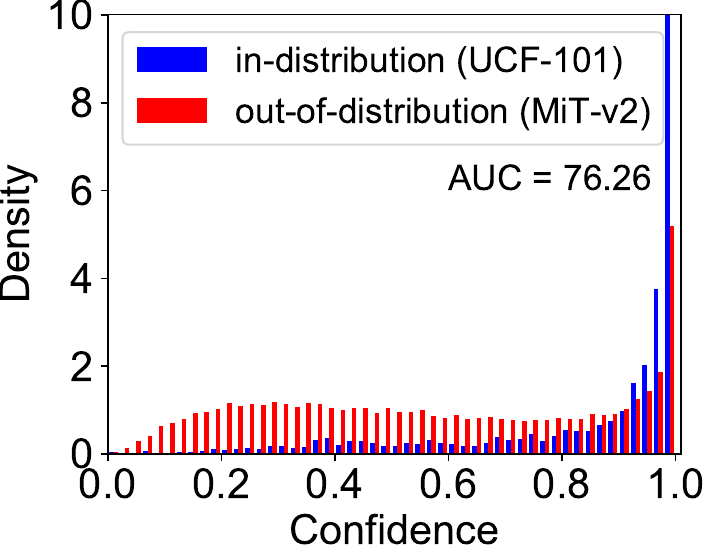}
    }
    \subcaptionbox{RPL~\cite{ChenECCV2020}}{
        \includegraphics[width=0.32\textwidth]{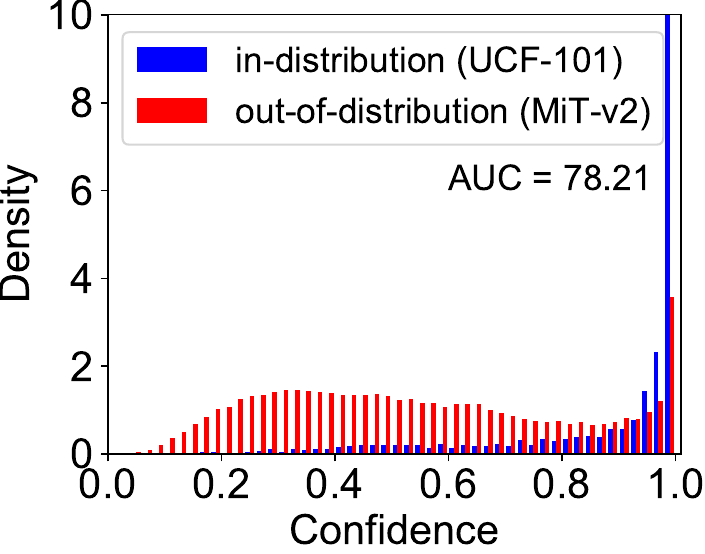}
    }
    \vfill
    \subcaptionbox{MC Dropout}{
        \includegraphics[width=0.32\textwidth]{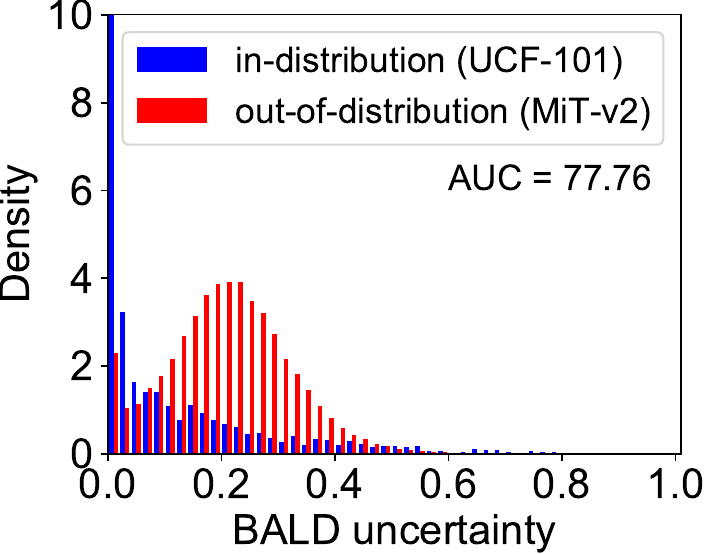}
    }
    \subcaptionbox{BNN SVI~\cite{KrishnanNIPS2018}}{
        \includegraphics[width=0.32\textwidth]{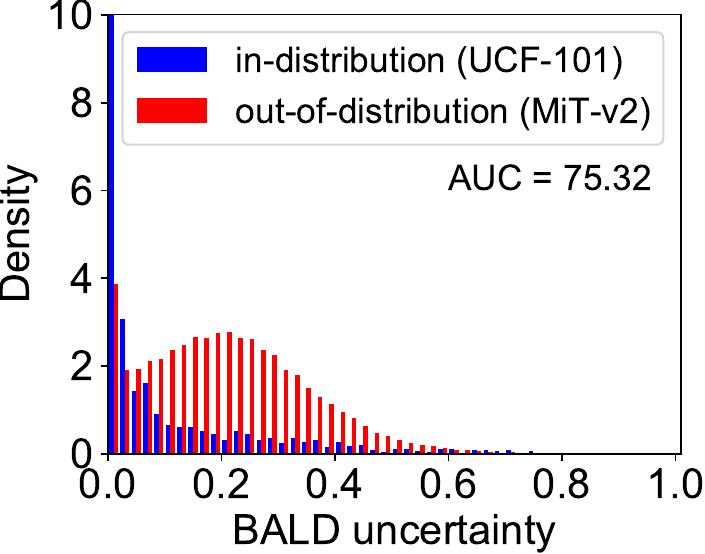}
    }
    \subcaptionbox{DEAR (full)}{
        \includegraphics[width=0.32\textwidth]{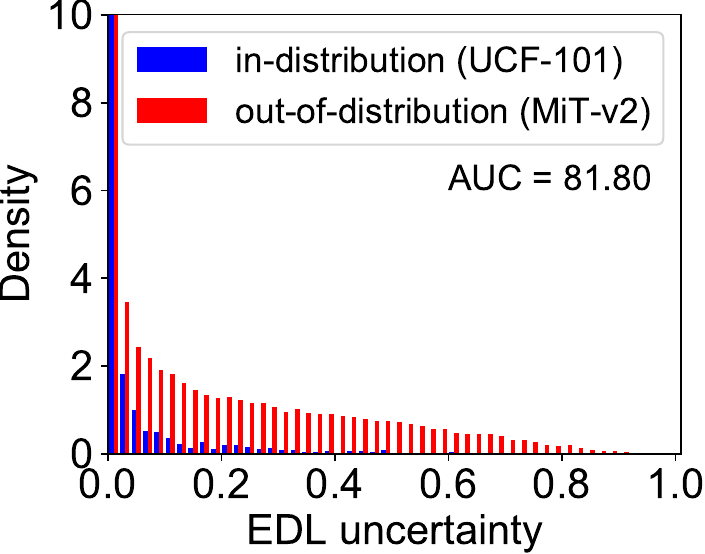}
    }
    \captionsetup{font=small,aboveskip=5pt}
    \caption{\textbf{TPN-based Out-of-distribution Detection with MiT-v2 as Unknown.} Values are normalized to [0,1] within each distribution.}
    \label{fig:ood_tpn_mit}
\end{figure*}

\textbf{Out-of-Distribution Detection}. From Fig.~\ref{fig:ood_i3d_hmdb} to Fig.~\ref{fig:ood_tpn_mit}, we provide the out-of-distribution detection results to compare our performance with all baselines listed in the main paper. Results on both HMDB-51 and MiT-v2 datasets with I3D, TSM, SlowFast, and TPN are provided. Note that OpenMax, SoftMax, and RPL are not predicting the uncertainty score of input sample, we instead use the confidence score (the maximum score of categorical probabilities) to show the OOD detection performance. These figures show that the uncertainties estimated by the proposed DEAR method exhibit a more long-tailed and flatten distribution than those estimated by MC Dropout and BNN SVI.

\section{Qualitative Results}
\label{quali}

\begin{figure*}[t]
    \centering
    \subcaptionbox{I3D}{
        \includegraphics[width=0.235\textwidth]{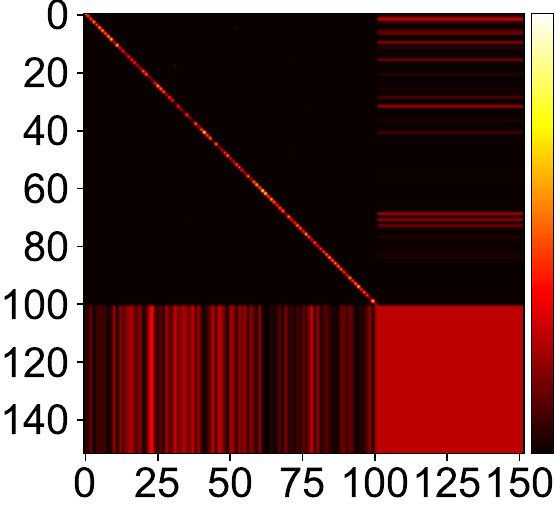}
    }
    \subcaptionbox{TSM}{
        \includegraphics[width=0.235\textwidth]{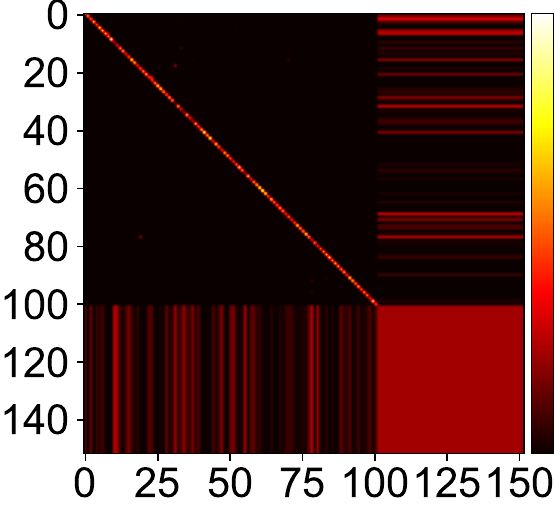}
    }
    \subcaptionbox{SlowFast}{
        \includegraphics[width=0.235\textwidth]{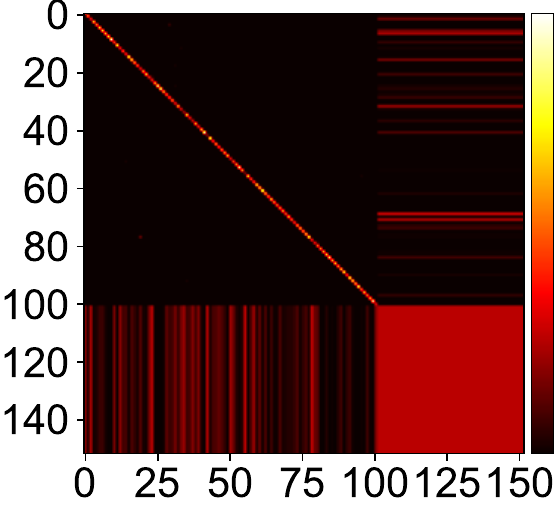}
    }
    \subcaptionbox{TPN}{
        \includegraphics[width=0.235\textwidth]{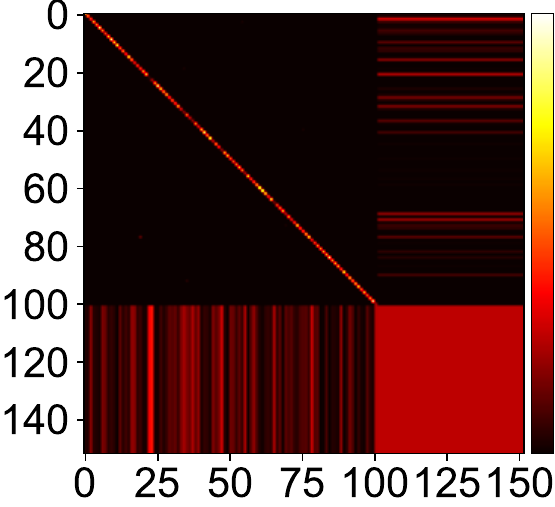}
    }
    \captionsetup{font=small,aboveskip=5pt}
    \caption{\textbf{Confusion Matrices of DEAR using HMDB-51 as Unknown.} The $x$-axis and $y$-axis represent the ground truth and predicted labels, respectively. The first 101 rows and columns are known classes from UCF-101 while the rest 51 classes are unknown from HMDB-51. Values are uniformly scaled into [0,1] and high value is represented by a lighter color (best viewed in color).}
    \label{fig:confmat_hmdb}
\end{figure*}
\begin{figure*}[t]
    \centering
    \subcaptionbox{I3D}{
        \includegraphics[width=0.235\textwidth]{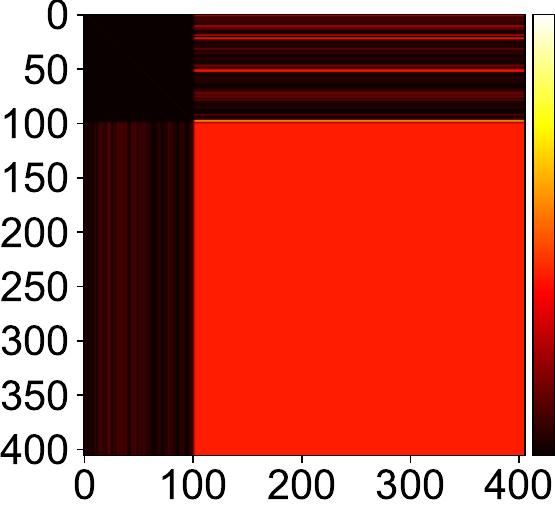}
    }
    \subcaptionbox{TSM}{
        \includegraphics[width=0.235\textwidth]{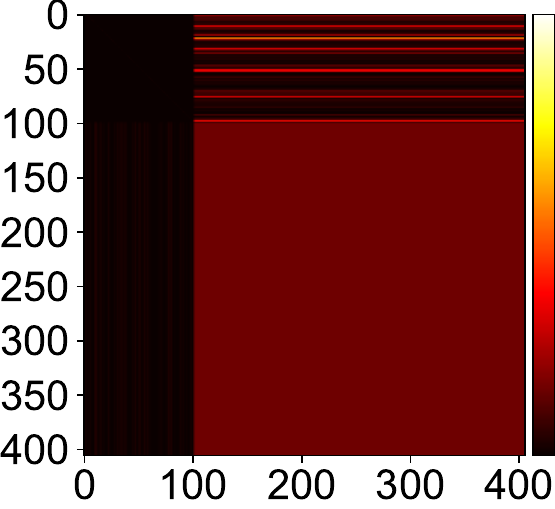}
    }
    \subcaptionbox{SlowFast}{
        \includegraphics[width=0.235\textwidth]{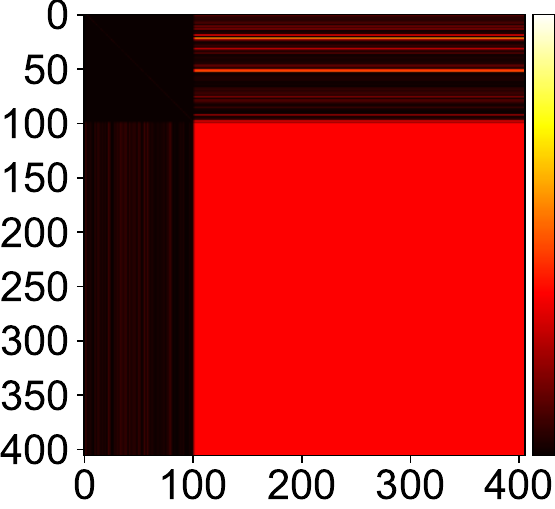}
    }
    \subcaptionbox{TPN}{
        \includegraphics[width=0.235\textwidth]{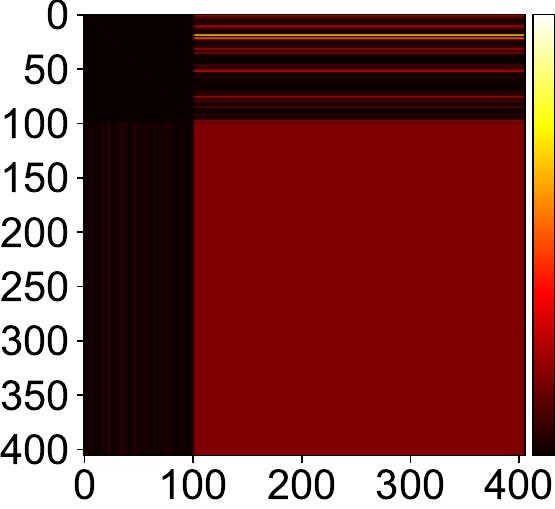}
    }
    \captionsetup{font=small,aboveskip=5pt}
    \caption{\textbf{Confusion Matrices of DEAR using MiT-v2 as Unknown.} The $x$-axis and $y$-axis represent the ground truth and predicted labels, respectively. The first 101 rows and columns are known classes from UCF-101 while the rest 305 classes are unknown from MiT-v2. Values are uniformly scaled into [0,1] and high value is represented by a lighter color (best viewed in color).}
    \label{fig:confmat_mit}
\end{figure*}

\textbf{Open Set Confusion Matrix}. In Fig.~\ref{fig:confmat_hmdb} and Fig.~\ref{fig:confmat_mit}, we provide the confusion matrix results. These figures show that when HMDB-51 dataset is used as the unknown, the ratio of mis-classification that classifying the samples from known classes into unknown (see the bottom-left region in each sub-figure) is less on TSM and SlowFast models than that on I3D and TPN models. When MiT-v2 dataset is used as the unknown, the unknown classes are the dominant testing case and from the bottom-right region, we see that the proposed method on I3D and SlowFast models shows significant advantage (brighter red color) over the method on TSM and TPN.


\textbf{Representation Debiasing Examples}. In Fig.~\ref{fig:eg_debias}, we provide examples of three classes, i.e., \textit{playing piano}, \textit{writing}, and \textit{golf driving} from both the biased dataset Kinetics and the unbiased (out-of-context) dataset Mimetics. We compare the recognition results of the variants of our proposed DEAR method with and without CED. These examples show that the CED module could help the DEAR method to recognize human actions on both the biased and unbiased datasets. For example, without the CED module, the model falsely recognizes the \textit{golf driving} as \textit{shooting soccer goal}. The reason could be conjectured that these video samples of the two classes are similar in the static background, i.e., large area of green grassland.


\clearpage




\newcommand{\figwidth}{0.21\textwidth}
\begin{figure*}[t]
\centering
\renewcommand{\tabcolsep}{2pt} %
\setlength{\extrarowheight}{0.5mm}
\begin{tabular}{l|c|c|c}
\hline
\parbox[c]{4mm}{\multirow{1}{*}[6.0em]{\makecell[l]{Kinetics \\ (Biased)}}} &
\includegraphics[width=\figwidth]{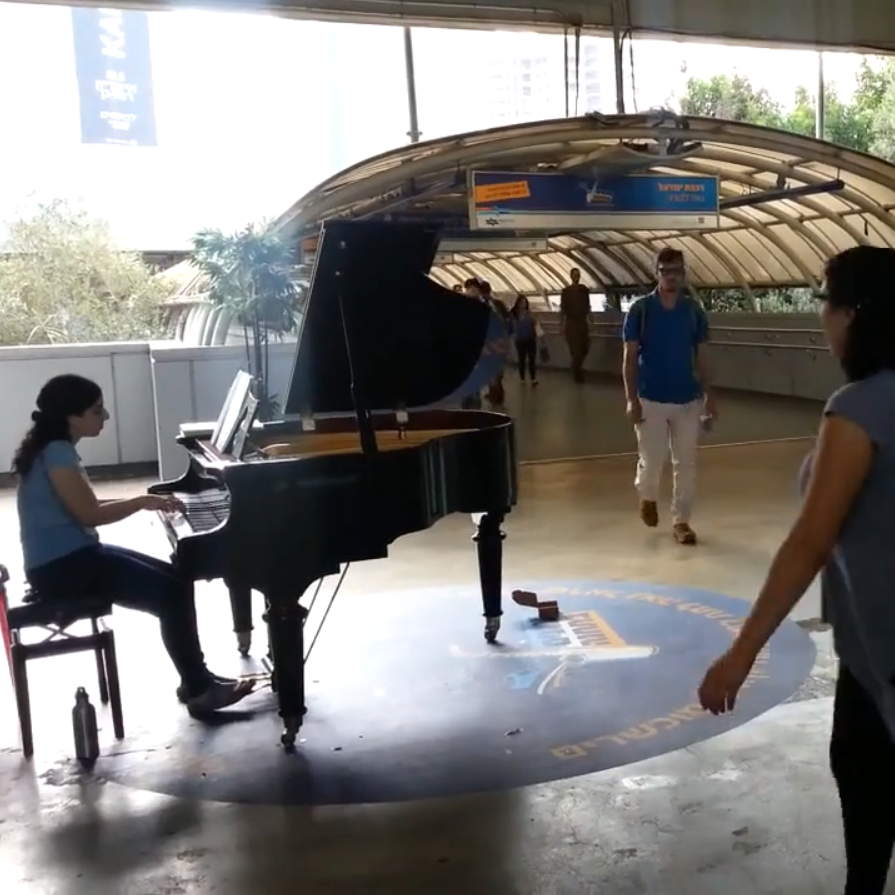} &
\includegraphics[width=\figwidth]{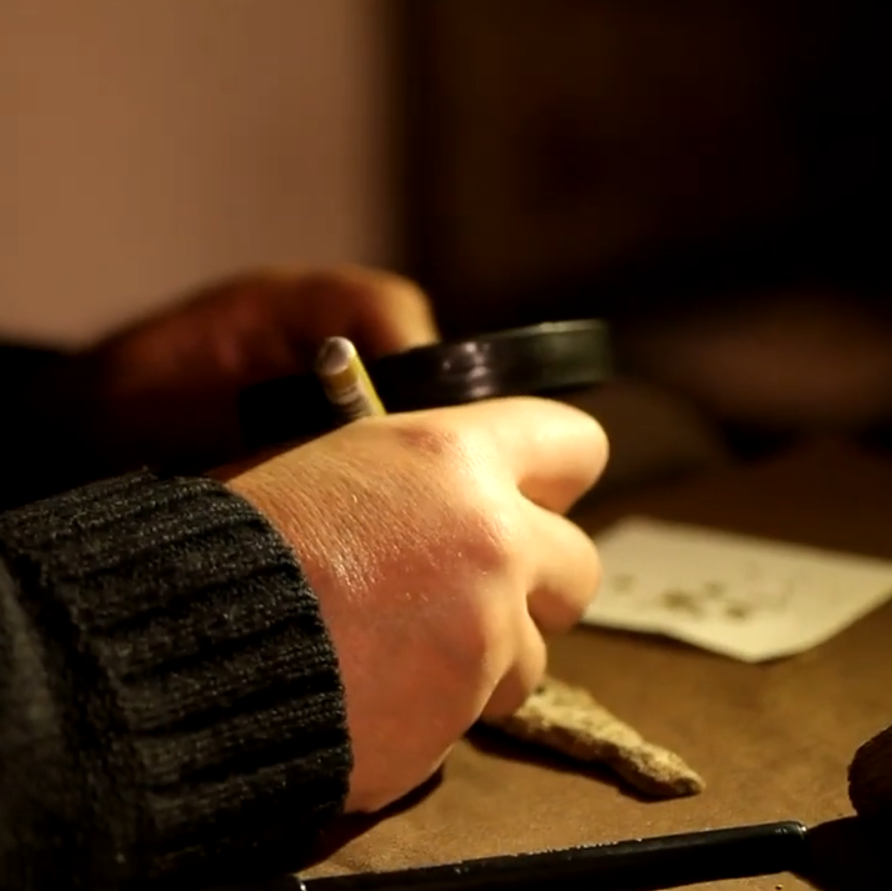} &
\includegraphics[width=\figwidth]{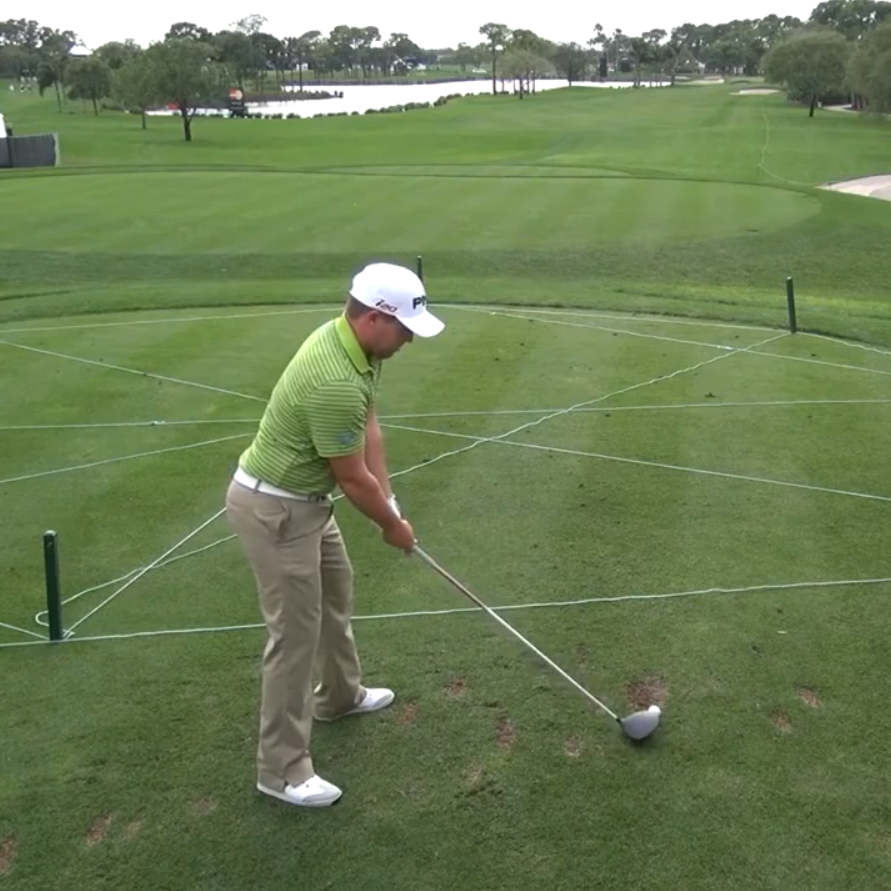} 
\\
\hline
DEAR (w/o CED) & Playing Volleyball (\xmark) & Opening Bottle (\xmark) & Shooting Soccer Goal (\xmark)
\\
DEAR (full) & Playing Piano (\checkmark) & Writing (\checkmark) &  Golf Driving (\checkmark) \\
\hline
\parbox[c]{4mm}{\multirow{1}{*}[6.0em]{\makecell[l]{Mimetics \\ (Unbiased)}}} &
\includegraphics[width=\figwidth]{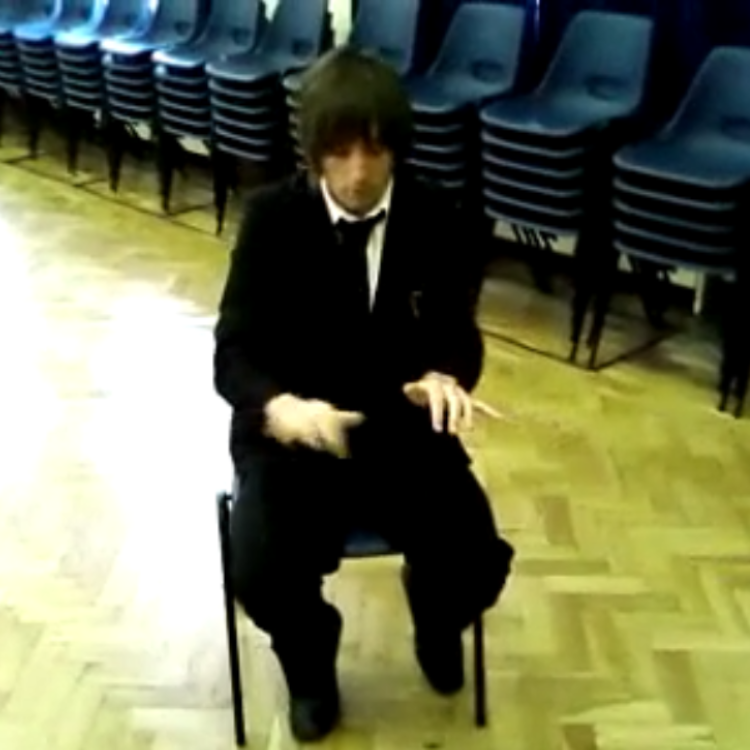} &
\includegraphics[width=\figwidth]{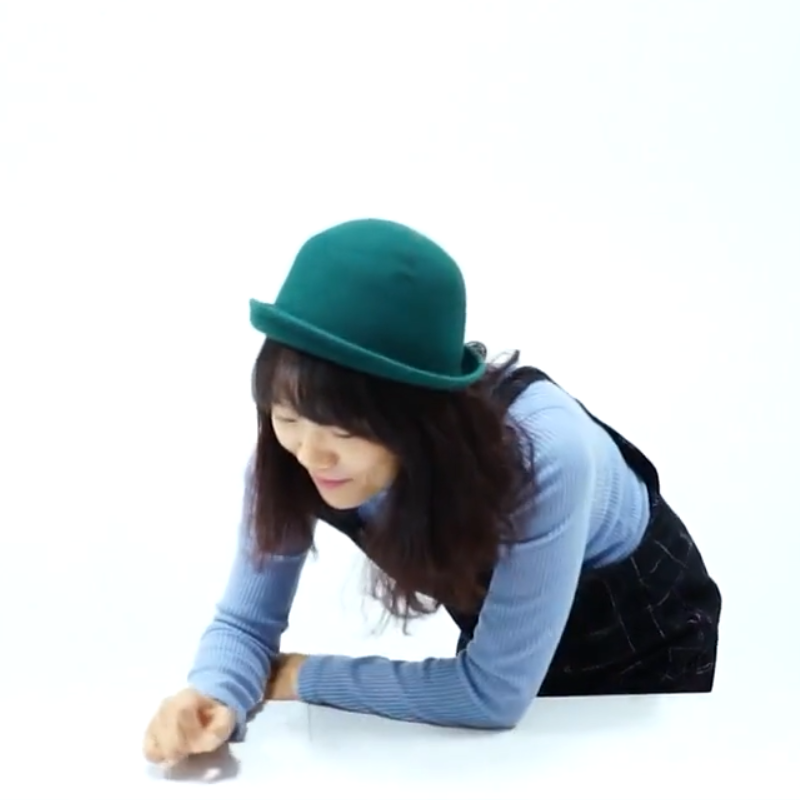} &
\includegraphics[width=\figwidth]{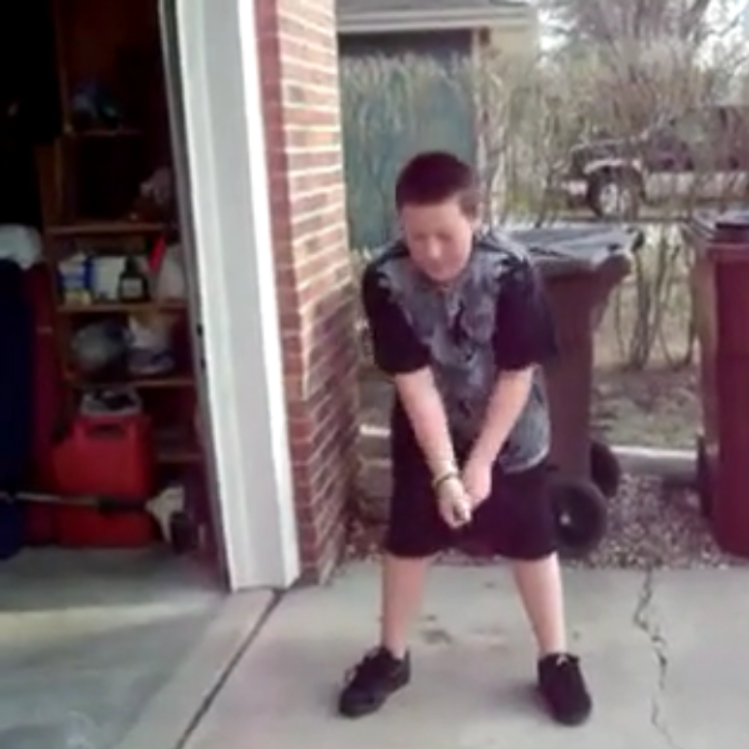} 
\\
\hline
DEAR (w/o CED) & Golf Driving (\xmark) & Golf Driving (\xmark) &  Opening Bottle (\xmark)
\\
DEAR (full) &  Playing Piano (\checkmark) &  Writing (\checkmark) &  Golf Driving (\checkmark) \\
\hline
\end{tabular}
\captionsetup{font=small,aboveskip=5pt}
\caption{\textbf{Examples of Kinetics and Mimetics.} The check mark (\checkmark) indicates that the predicted label is correct while the cross mark (\xmark) means that the predicted label is incorrect.}
\label{fig:eg_debias}
\end{figure*}

\end{appendix}

\end{document}